\documentclass{article}          
\usepackage{arxiv}
\usepackage[utf8]{inputenc} 
\usepackage[T1]{fontenc}    
\usepackage{hyperref}       
\usepackage{url}            
\usepackage{booktabs}       
\usepackage{amsfonts}       
\usepackage{nicefrac}       
\usepackage{microtype}      
\usepackage{graphicx}
\usepackage{tabulary}
\usepackage{multirow} 
\usepackage{lscape}
\usepackage{lmodern}
\usepackage{amsmath}
\usepackage{rotating}
\usepackage{algorithm}
\usepackage{algorithmic}
\usepackage{tikz}
\usepackage{amsmath}
\usepackage{tikz}
\usetikzlibrary{arrows,shapes,shadings,patterns}
\usetikzlibrary{positioning}
\usepackage{pgfplots}
\usepackage{float}
\usepackage{bm}
\usepackage{stfloats}
\newcolumntype{K}[1]{>{\centering\arraybackslash}p{#1}}

\begin{document}

\title{Multi-fidelity modeling with different input domain definitions using Deep Gaussian Processes}

\author{%
  Ali Hebbal \thanks{Ph.D. student, ONERA, DTIS, Universit\'e Paris Saclay, Universit\'e de Lille, CNRS/CRIStAL, Inria Lille, ali.hebbal@onera.fr )} \\
  ONERA, Inria, France
\\
  \texttt{ali.hebbal@onera.fr}  \\
   \And
   Loic Brevault \thanks{Research Engineer, ONERA, DTIS, Universit\'e Paris Saclay, loic.brevault@onera.fr} \\
   ONERA, France\\
   \texttt{loic.brevault@onera.fr} \\
   \And
   Mathieu Balesdent \thanks{Research Engineer, ONERA, DTIS, Universit\'e Paris Saclay, mathieu.balesdent@onera.fr  } \\
   ONERA, France\\
   \texttt{mathieu.balesdent@onera.fr} \\
   \And
   El-Ghazali Talbi \thanks{
Professor at Polytech’Lille - Universit\'e de Lille, el-ghazali.talbi@univ-lille1.fr} \\
   Inria Lille, France\\
   \texttt{el-ghazali.talbi@univ-lille.fr}
   \And
   Nouredine Melab \thanks{
Professor at Polytech’Lille - Universit\'e de Lille, el-ghazali.talbi@univ-lille1.fr} \\
   Inria Lille, France \\
   \texttt{nouredine.melab@univ-lille.fr} \\
}

\maketitle

\begin{abstract}

Multi-fidelity approaches combine different models built on a scarce but accurate data-set (high-fidelity data-set), and a large but approximate one (low-fidelity data-set) in order to improve the prediction accuracy. Gaussian Processes (GPs) are one of the popular approaches to exhibit the correlations between these different fidelity levels. Deep Gaussian Processes (DGPs) that are functional compositions of GPs have also been adapted to multi-fidelity using the Multi-Fidelity Deep Gaussian process model (MF-DGP). This model increases the expressive power compared to GPs by considering non-linear correlations between fidelities within a Bayesian framework. However, these multi-fidelity methods consider only the case where the inputs of the different fidelity models are defined over the same domain of definition (\textit{e.g.}, same variables, same dimensions).  However, due to simplification in the modeling of the low-fidelity, some variables may be omitted or a different parametrization may be used compared to the high-fidelity model. In this paper, Deep Gaussian Processes for multi-fidelity (MF-DGP) are extended to the case where a different parametrization is used for each fidelity. The performance of the proposed multi-fidelity modeling technique is assessed on analytical test cases and on structural and aerodynamic real physical problems.
\end{abstract}

\noindent\textbf{Notations:}
\begin{itemize}
\item A scalar is represented by a lower case character: $y\in \mathbb{R}$
\item A vector is represented by a bold character:  $\textbf{x}\in \mathbb{R}^d, \textbf{x}=[x_1,\ldots,x_d]^\top$
\item A matrix is represented by upper case character: $X=\begin{bmatrix} x_{1,1} &\cdots&x_{1,j} &\cdots & x_{1,d} \\\vdots &\ddots &\vdots &\ddots& \vdots \\x_{i,1} &\cdots&x_{i,j} &\cdots & x_{i,d}\\\vdots &\ddots &\vdots &\ddots& \vdots\\x_{n,1} &\cdots&x_{n,j} &\cdots & x_{n,d} \end{bmatrix} \in \mathbb{R}^{n\times d} $
\item The $i^{th}$ row of a matrix X is noted $\textbf{x}^{(i)\top} \in \mathbb{R}^d $ 
\item The $j^{th}$ column of a matrix X is noted $\textbf{x}_j  \in \mathbb{R}^n $ 
\end{itemize}

\section{Introduction}

The analysis of complex systems often involves high-fidelity simulation codes that require computationally intensive evaluations to assess the response of interest. Often, for the analysis of these high-fidelity functions and for on-line decision making, a surrogate model is constructed from a small high-fidelity (HF) data-set \cite{wang2007review}, \cite{forrester2008engineering}. However, due to the limited size of this data-set, the quality of a surrogate model prediction based only on the high-fidelity data is usually of poor quality. Multi-fidelity approaches \cite{fernandez2016review}, \cite{peherstorfer2018survey} are used to overcome this issue by enhancing the high-fidelity data with low-fidelity (LF) model evaluations that are computationally cheaper to obtain but are less accurate. This can be accomplished by three main modeling approaches:
\begin{itemize}
\item numerical relaxation, for instance, in a simulation code that requires an optimization sub-problem to be solved, a low number of iterations in the optimization process is chosen for the low-fidelity model. In \cite{jonsson2015shape}, for the shape optimization of trawl-doors a low-fidelity CFD model similar to the high-fidelity CFD model is used but with a relaxed flow solver convergence criteria, this results in a LF model 78 times faster than the HF model.
\item different assumptions about the physical model by neglecting some physical effects. For instance, in \cite{iyappan2020multi} an Euler-Bernoulli beam finite element model \cite{reddy1993introduction} is considered as the low-fidelity to compute the load-carrying and deflection characteristics of a short beam, the effects of rotary inertia and shear deformation are neglected in this model and the cross-section remains perpendicular to the bending axis. Whilst in the high-fidelity the Timoshenko beam theory \cite{reddy1993introduction} is used, which takes into accounts the effects of rotary inertia and shear deformation and the cross-section has no longer to be perpendicular to the bending axis for short and small beams.
\item different levels of space or time discretization. For example, in \cite{brooks2011multi}, for the aerodynamic shape optimization of a transonic compressor rotor, in the low-fidelity a coarse mesh refinement is used to solve the Reynolds-Averaged steady Navier-Stokes (240000 nodes), whilst in the high-fidelity, a fine mesh grid is used (740000 nodes).
\end{itemize}

\begin{figure*}[b]
\begin{minipage}[c]{0.48\linewidth} 
\center
\includegraphics[width=1.\linewidth]{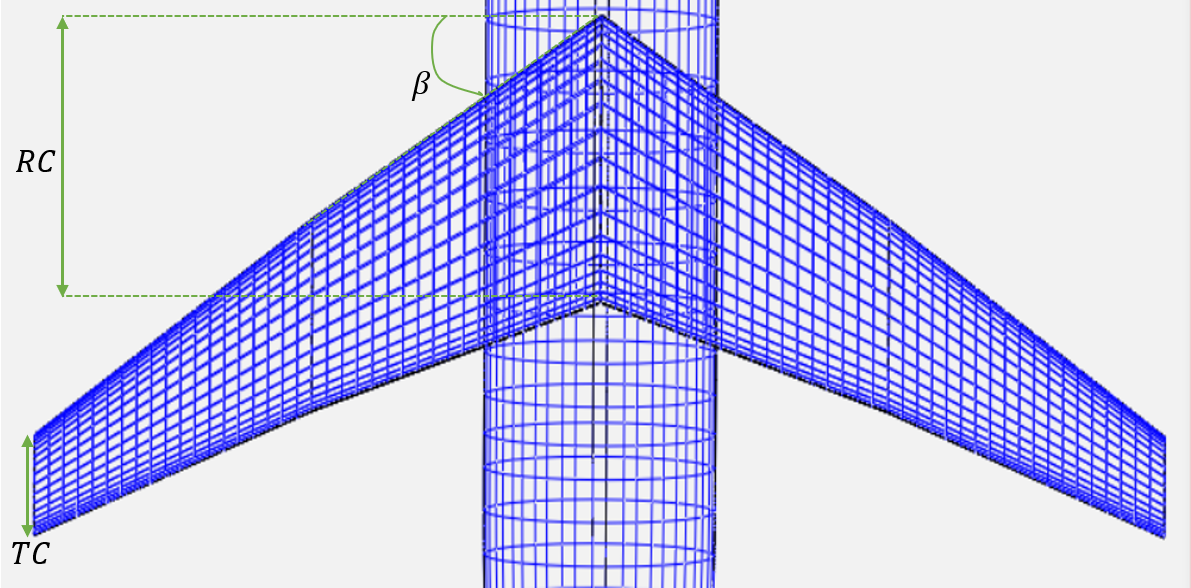}
\end{minipage}
\hfill
\begin{minipage}[c]{0.48\linewidth} 
\includegraphics[width=1.\linewidth]{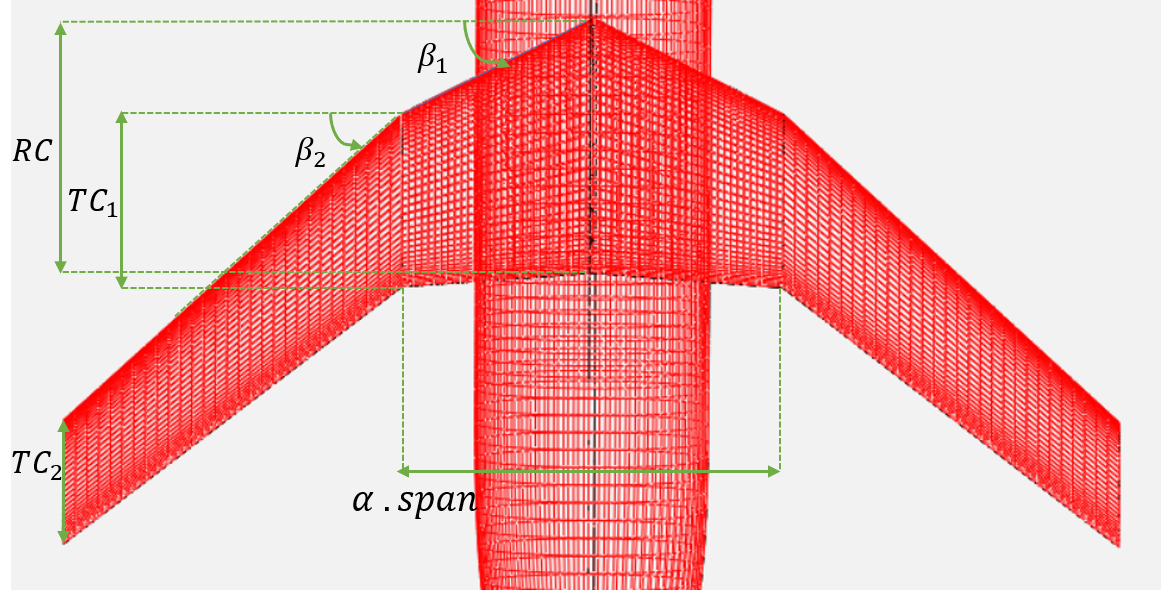}
\end{minipage}
\caption{ A one-section wing characterized by 3 design variables: its root chord ($RC$), tip chord ($TC$), and the sweep angle ($\beta$) (left) can be used as a low-fidelity model of a two-section wing characterized by 6 design variables: its root chord ($RC$), tip chord of the first section ($TC_1$), tip chord of the second section ($TC_2$), sweep angle of the first section ($\beta_1$), sweep angle of the second section ($\beta_2$) and the relative span of the first section ($\alpha$) (right).}
\label{sections_hf_lf}
\end{figure*}
Multi-fidelity modeling has been a popular research topic both in the engineering and machine learning communities. In fact, different models have been developed based on Gaussian processes \cite{kennedy2001bayesian}, \cite{le2014recursive},  \cite{raissi2016deep}, \cite{perdikaris2017nonlinear}, \cite{cutajar2019deep}, neural networks \cite{kim2007hybrid}, \cite{minisci2013robust} or support vector machines \cite{shi2019support} and applied to a large spectrum of engineering applications including aerodynamics \cite{kuya2011multifidelity}, \cite{shah2015multi}, electronics \cite{bekasiewicz2015efficient}, thermodynamics \cite{reeve2017error}, or mechanics \cite{vitali2002multi}. However, a rarely investigated case is where the input space definition is different in each fidelity. In fact, in practice for the sake of simplicity, the LF model may not consider some input variables. For instance, in aerodynamics, to model multiple-section wing, simplified planform characterization can be used considering one section with average chords and sweep angles (Fig.~\ref{sections_hf_lf}). Moreover, different parametrizations can be used in each fidelity. For instance, for geometrical input variables in one fidelity, a Cartesian formulation can be used, whilst in the other fidelity, a spherical formulation is preferred. The classical approaches in literature for multi-fidelity modeling with different parametrizations of the input spaces are based on space mapping \cite{bandler2004space}. The basic idea consists in finding a parametric mapping from the input space of the high-fidelity to the input space of the low-fidelity which minimizes a defined distance between the responses of the two models. These approaches have been intensively used for multi-fidelity optimization \cite{bandler2004space}, \cite{bandler2006space}, \cite{robinson2008surrogate}, \cite{koziel2010computationally} where the mapping is done locally using trust-region algorithms around the local optimum. These local approaches are suited for the optimization task but may not be adapted for modeling on the entire input space since the mapping is performed around the optimum. A dedicated modeling approach has been developed \cite{tao2019input} which consists of an input mapping calibration (IMC) for the entire definition domain of the input space. However, the input mapping calibration is based on the concept that the low-fidelity model has similar trend as the high-fidelity one.

In this paper, a new model based on the multi-fidelity Deep Gaussian process model (MF-DGP) \cite{cutajar2019deep} is proposed for multi-fidelity problems with different input spaces. This is accomplished by a new model formulation of MF-DGP incorporating the mapping between different fidelity input spaces in a non-parametric way, based only on the nominal values of the mapping of the input HF training data. 

The contribution of this paper is twofold. First, a new model based on MF-DGP is proposed for multi-fidelity modeling with different input space parametrizations. This model uses a non-parametric and non-deterministic input mapping, and performs a joint optimization of the multi-fidelity model and the input mapping. Moreover, the proposed model is adapted for the context when the nominal mapping is computationally expensive to obtain or is known only for the training HF data. Second, the model is assessed on analytical test cases and physical test problems and compared to existing approaches.

The rest of the paper is organized into three main sections. Section 2 provides a review of literature on the different multi-fidelity models based on Gaussian Processes and on the approaches of space mapping to handle the variable parametrization of the input spaces. In Section 3, the proposed model is described. Finally, Section 4 presents experimentations on analytical test problems and on real physical test problems (structural and aerodynamical test cases) to assess the performance of the proposed model.

\section{Background}
\subsection{Multi-fidelity using Gaussian Processes}
One of the most popular multi-fidelity modeling approaches is based on Gaussian Processes (GPs) \cite{rasmussen2006gaussian}. The interesting feature of GPs is their Bayesian formulation which induces uncertainty quantification in addition to prediction capability. The proposed model is derived from GPs, thus, a brief introduction to GPs and to multi-fidelity models based on GPs is presented in this Section.
\subsubsection{Gaussian Processes}
Gaussian Processes (GP) \cite{rasmussen2006gaussian} are a popular approach for regression problems. Given a Design of Experiments (DoE) $X = \left(x_j^{(i)}\right)_{{1\leq i \leq n}\atop 1\leq j \leq d}$ and its associated scalar response values $\textbf{y}= \left(y^{(i)}\right)_{1\leq i \leq n}$, with $d$ the dimension of the input variables and $n$ the number of observations, a GP $f(\cdot)$ maps between the considered inputs/outputs: $$\textbf{y}=f(X) + \epsilon$$
where $\epsilon$ is a zero-mean Gaussian noise with a variance $\sigma^2$ used to account for noisy observations. $f(\cdot)$ is completely defined by its mean function $\mu(\cdot)$ (that is usually considered equal to a constant $\mu$ \cite{jones1998efficient}) and a parametrized covariance function $k^{\boldsymbol{\theta}}(\cdot,\cdot)$. The parameters $\boldsymbol{\theta}$ of the covariance function, the value of the constant mean function $\mu$, and the variance $\sigma^2$ of the Gaussian noise can be estimated using a maximum likelihood procedure \cite{rasmussen2006gaussian} 
\begin{equation}
\hat{\boldsymbol{\theta}}, \hat{\sigma}, \hat{\mu} = \underset{\boldsymbol{\theta}, \sigma, \mu}{\text{argmax}} \log(p(\textbf{y}|X))
 \end{equation}
 with
\begin{equation}
\begin{split} 
 \log(p(\textbf{y}|X))\propto& - \textbf{y}^T\left(k^{\boldsymbol{\theta}}(X,X)+\sigma^2 I_n\right)\textbf{y}\\& - \log\left|k^{\boldsymbol{\theta}}(X,X)+\sigma^2 I_n\right| - n\log(2\pi)
\end{split}
 \end{equation}
 where $p(\textbf{y}|X)$ is the marginal likelihood of the observed data and  $I_n $ is the identity matrix of size $n$. 
 
For the prediction task, at an unobserved point $\textbf{x}^*$, the prior distribution $f(\textbf{x}^*)$ is conditioned on the existing observations $f(X) = \textbf{y}$:
\begin{equation}
f(\textbf{x}^*)|f(X)\sim \mathcal{N}\left( \hat{f}(\textbf{x}^*), \hat{s}(\textbf{x}^*)^2 \right)
\end{equation}
where $\hat{f}(\textbf{x}^*)$ and $\hat{s}(\textbf{x}^*)^2$ are respectively the mean and the variance of the posterior distribution and are defined as:
\begin{equation}
\hat{f}(\textbf{x}^*)=\hat{\mu}+k^{\hat{\boldsymbol{\theta}}}(\textbf{x}^*,X){\left(k^{\hat{\boldsymbol{\theta}}}(X,X)+\hat{\sigma}^2I_n\right)}^{-1}(\textbf{y}-\hat{{\mu}})
\end{equation}
\begin{equation}
\hat{s}(\textbf{x}^*)^2 =  k^{\hat{\boldsymbol{\theta}}}(\textbf{x}^*,\textbf{x}^*)-k^{\hat{\boldsymbol{\theta}}}(\textbf{x}^*,X){\left(k^{\hat{\boldsymbol{\theta}}}(X,X)+\hat{\sigma}^2I_n \right)}^{-1}k^{\hat{\boldsymbol{\theta}}}(X,\textbf{x}^*)
\end{equation}

\subsubsection{Multi-fidelity models}
Let $(X^t,\textbf{y}^t)$ be the couple of inputs/outputs of each fidelity $t \in \{1,\ldots,s\}$, where $s$ is the number of fidelities. Let $d$ and $n_t$ be respectively the dimension of the input data and the size of the training data at fidelity $t$.

Due to their attractive features, GPs have been extended to multi-fidelity modeling. One of the most popular multi-fidelity based GP approaches is the Auto-Regressive (AR1) model  \cite{kennedy2001bayesian}, \cite{le2014recursive}. It is based on a linear correction between the high and low-fidelity outputs. More specifically, a GP prior is assigned to each fidelity $t$, where the HF prior $f_t(\cdot)$ is equal to the LF prior $f_{t-1}(\cdot)$ multiplied by a scaling factor $\rho$ plus an additive bias GP $\gamma_t(\cdot)$:
\begin{equation}
f_t(\textbf{x}) = \rho f_{t-1}(\textbf{x}) + \gamma_t(\textbf{x}), \forall \textbf{x}\in \mathbb{R}^d
\end{equation}

Two formulations of AR1 may be distinguised: a fully coupled formulation \cite{kennedy2001bayesian} and a recursive formulation \cite{le2014recursive}. The latter supposes a nested structure of the DoE and enables a reduction of the computational complexity of the model training from $\mathcal{O} \left(\sum_{t=1}^{s} n_t\right)^3 $ to $\mathcal{O}\left( \sum_{t=1}^{s} \left(n_t\right)^3 \right) $ where $s$ is the number of fidelities and $n_t$ is the number of observations at fidelity $t$.

\begin{figure}[b]%
\centering
\input{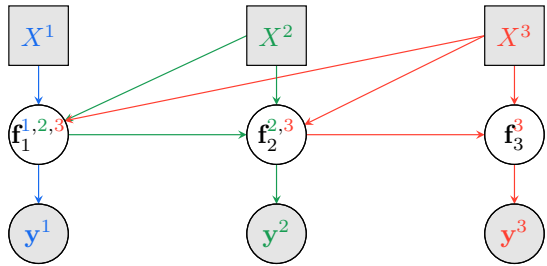}%
\caption{Graphical representation of MF-DGP model \cite{cutajar2019deep}. Each layer of the DGP corresponds to a fidelity.}
\label{MF-DGP}
\end{figure}

AR1 assumes a linear relationship between the fidelities. However, it happens that the correlation between fidelities is non-linear and a linear correction is not adapted \cite{perdikaris2017nonlinear}. A more global approach considers the HF prior equal to a non-linear transformation of the LF prior by a GP prior $h_{t}(\cdot)$ plus an independent additive bias GP:
\begin{equation}
f_t(\textbf{x}) = h_t \left( f_{t-1}(\textbf{x})\right) +  \gamma_t(\textbf{x}), \forall \textbf{x}\in \mathbb{R}^d
\label{non_linear}
 \end{equation}
This equation has given rise to two approaches: the Non-linear Auto-Regressive multi-fidelity GP model (NARGP) \cite{perdikaris2017nonlinear} and the Multi-Fidelity Deep Gaussian Process model (MF-DGP) \cite{cutajar2019deep}. The NARGP relaxes Eq.~\ref{non_linear} by considering each fidelity as a transformation of the GP posterior of the lower-fidelity instead of the prior, which enables the GPs to be trained sequentially under the hypothesis of a nested fidelity structure of the DoE as in the recursive AR1. MF-DGP, on the other hand, keeps the exact relationship, which comes back to a Deep Gaussian Process (DGP) \cite{damianou2013deep} where each layer corresponds to a fidelity level (Fig.~\ref{MF-DGP}). MF-DGP is based on the sparse DGP approximation proposed in \cite{salimbeni2017doubly} which consists in introducing a set of inducing inputs/outputs at each layer then approximating the distribution of the inducing outputs using a variational approximation. However, the training of MF-DGP is more difficult than regular DGPs due to the optimization of the induced inputs in an augmented input space (see Section 3.4).

%

\subsubsection{Variable input space parametrization}
The majority of multi-fidelity approaches assume that fidelities share the same input space. However, in practice, this is not always the case. In fact, due to either different modeling approaches from one fidelity to another, or omission of some variables in the lower-fidelity models, the input spaces may have distinct parametrization forms and/or dimensionality.

In the literature, the main multi-fidelity approaches that address this issue belong to the space mapping multi-fidelity class \cite{bandler2004space}. The space mapping multi-fidelity methods act on the inputs rather than the outputs of the models. The basic concept is to transform the high-fidelity inputs using a parametric function in order to minimize a distance between the corresponding low-fidelity outputs of this mapping and the exact high-fidelity outputs. In the space mapping approaches two fidelities are considered, hence, instead of using the $s$ levels of fidelity notation, the couple of high-fidelity and low-fidelity inputs/outputs data are respectively noted $(X^{hf},\textbf{y}^{hf})$ and $(X^{lf},\textbf{y}^{lf})$ for the space mapping approaches:
\begin{equation}
\hat{\beta} = \underset{\boldsymbol{\beta}}{\text{argmin}} \sum_{i=1}^k\left(||{y^{hf}}^{(i)}-f_{lf}^{\text{exact}}\left(g_{\boldsymbol{\beta}}({{\textbf{x}}^{hf}}^{(i)})\right)||\right)
\end{equation}
where $k$ denotes the size of the set of mapped points which is a subset of the HF data chosen based on trust region optimization algorithms and  $g_{\boldsymbol{\beta}}(\cdot)$ corresponds to the mapping and $\boldsymbol{\beta}$ to its vector of parameters (in most applications $g_{\boldsymbol{\beta}}(\cdot)$  is considered linear). The space mapping has been extensively used for multi-fidelity \cite{bandler2004space},  \cite{bandler2006space}, \cite{robinson2008surrogate}, \cite{koziel2010computationally} and different parametric mappings have been used, for instance, aggresive space mapping \cite{rayas2016power} and neural networks \cite{rayas2004based}.

It was first used in the case of variable-size input parametrization in \cite{robinson2008surrogate}. However, the space mapping approaches are used in an optimization context, and the mapping is used around the optimum candidates and is updated at each iteration of the trust-region optimization. This is not suited from a modeling point of view where the analysis of the high-fidelity function is performed for the whole input space.

A nominal mapping $g_0(\cdot)$, based on practical insights of the multi-fidelity problem, is usually required. It expresses the assumed relationship between the different input spaces. In some cases, this nominal mapping is trivial. For instance, if the set of high and low-fidelity inputs are from the same set of physical equations, the low-fidelity inputs can then be a subset
of the high-fidelity ones. Usually, the nominal mapping is problem specific and is defined based on expert opinion. A multi-fidelity approach as a bias correction \cite{li2016integrating} (BC) can then be used based on this nominal mapping:

\begin{equation}
f_{hf}(\textbf{x}) = f_{lf}(g_0(\textbf{x})) + \gamma(\textbf{x}), \forall \textbf{x}\in \mathbb{R}^d
\label{eq:BC}
\end{equation}

\begin{figure}[t]
\center
\input{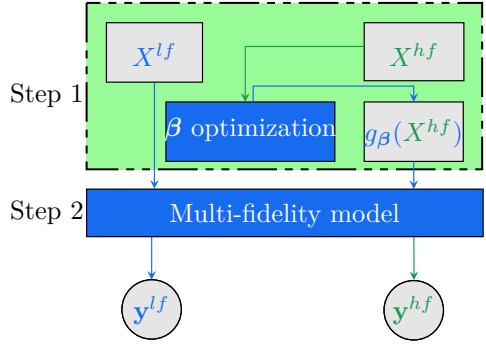}%
\caption{Graphical representation of the IMC approach.}
\label{IMC}
\end{figure}

The Input Mapping Calibration (IMC) \cite{tao2019input} is an approach that seeks to obtain a potentially better mapping than the nominal mapping. As the space mapping approach, it consists in finding a parametric mapping $g_{\boldsymbol{\beta}}(\cdot)$. However, here the mapping is considered for the whole input space and the parameters of the mapping are obtained by minimizing the difference between the LF and HF model outputs on the HF data points plus a regularization term $R(\boldsymbol{\beta},\boldsymbol{\beta}_0)$ based on the nominal mapping parameters $\boldsymbol{\beta}_0$:
\begin{equation}
\hat{\boldsymbol{\beta}} = \underset{\boldsymbol{\beta}}{\text{argmin}}  (\sum\limits_{i=1}^{n_{hf}} \left({y^{hf}}^{(i)}-f_{lf}^{\text{exact}}\left(g_{\boldsymbol{\beta}}({{\textbf{x}}^{hf}}^{(i)})\right)\right)^2  + R(\boldsymbol{\beta},\boldsymbol{\beta_0})) 
\label{eq:IMC}
\end{equation}
where $n_hf$ corresponds to the number of HF training data points. The high-fidelity input data is then projected with the obtained mapping on the low-fidelity input space, and a multi-fidelity model with the same input spaces can be used (Fig.~\ref{IMC}). This optimization of the mapping parameters is done previously to the training of the multi-fidelity model, which prevents the parameters of the mapping to be updated, once the multi-fidelity model is optimized. Besides, the optimization is done using the exact low-fidelity model, which is considered as computationally free to evaluate, however, in many applications, it may not be the case. Moreover, the correlations over the original HF input space are not taken into account, since the multi-fidelity model is trained only on the lower-fidelity input space. Finally, the mapping parameters are estimated based on the concept that the low-fidelity model shares a similar trend with the high-fidelity one. This is the case in some applications as microwave applications where space mapping has emerged. However, in many multi-fidelity problems, minimizing the distance between the outputs does not guarantee an appropriate mapping. These aspects are illustrated by an illustrative example in the first experiment of Section 4.  

Up until now, the mapping from the high to the low-fidelity inputs is based on parametric deterministic functions and is usually trained sequentially with the multi-fidelity model. The approach that is proposed in the next section is based on a non-parametric Bayesian mapping which is learned and is embedded in the multi-fidelity model.

\section{Proposed model}
In \cite{hebbal2019multi}, a first version of MF-DGP for varying input space parametrizations is developed. In this Section, an improved version named Multi-Fidelity Deep Gaussian Process Embedded Mapping (MF-DGP-EM) is presented with more details on its specifications.
\subsection{Description of MF-DGP-EM}

Let $(X^t,\textbf{y}^t)$ be the couple of inputs/outputs of each fidelity $t \in \{1,\ldots,s\}$, where $s$ is the number of fidelities. Let $d_t$ and $n_t$ be respectively the dimension of the input data and the size of the training data at fidelity $t$, and let $X_{t-1}^t$ be the nominal mapped values of the input data $X^t$ to the lower-fidelity $t-1$. 

The proposed model is based on the MF-DGP model presented in \cite{cutajar2019deep}. MF-DGP is applicable to the case of the same input space domain definition for all the fidelities. More specifically, a DGP is considered where each layer corresponds to a fidelity (Fig.~\ref{MF-DGP}). Moreover, the GP at each layer $t$ depends not only on the input data at this fidelity $X^t$ but also on all the previous fidelity evaluations for the same input data. To this end, $\textbf{f}^t_l$ denotes the evaluation at the layer $l$ of $X^t$, the input data at the fidelity $t$ (Fig.~\ref{MF-DGP}). This formulation of DGPs imposes the definition of a combination of covariance functions at each layer taking into account the correlation between the inputs as well as the correlation between the outputs. Hence, for two input data $\textbf{x}^{(i)}$ and $\textbf{x}^{(j)}$ the covariance at the fidelity level $t$ is defined as:
\begin{equation}
k_l(\textbf{x}^{(i)},\textbf{x}^{(j)}) = k_l^\rho\left(\textbf{x}^{(i)},\textbf{x}^{(j)}\right)  \times k_l^{f-1}\left(f^*_{l-1}\left(\textbf{x}^{(i)}\right),f^*_{l-1}\left(\textbf{x}^{(j)}\right)\right)+ k_l^\gamma\left(\textbf{x}^{(i)},\textbf{x}^{(j)}\right)
\nonumber
\end{equation}
where $f^*_{l-1}(\cdot)$ denotes the posterior of the GP at the layer $l-1$, $k_l^\rho(\cdot,\cdot)$ and $k_l^\gamma(\cdot,\cdot)$ are covariance functions with respectively an input space-dependent scaling effect and an input space-dependent bias effect, whilst $k_l^{f-1}(\cdot,\cdot)$ is the covariance function between the evaluated outputs at the previous layer.     

\begin{figure}
\centering
\input{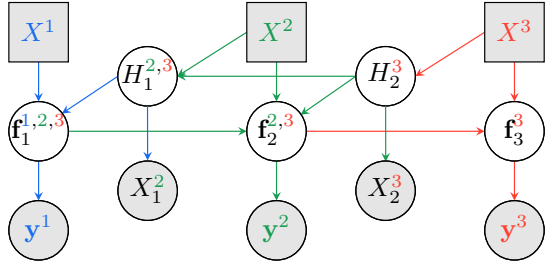}
\caption{Graphical representation of MF-DGP-EM with different input spaces.}
\label{newmodel}
\end{figure}

However, since each fidelity is defined on its own input space, MF-DGP can not be used directly in the case of different input space parametrizations. To overcome this issue, multi-outputs GPs $\left(H_t\right(\cdot))_{1\leq t\leq s-1}$ are introduced to map between the input spaces of two successive fidelities $t$ and $t+1$ . The input mapping GPs are conditioned on the nominal values of the input training HF data (Fig.~\ref{newmodel}). The model obtained is a two-level DGP, where the first level maps between the different fidelity input spaces and the second level propagates the fidelity evaluations. Hence, the mapping between the input spaces of the fidelities is defined within the multi-fidelity model.

This proposed model allows a concurrent optimization of the mapping and the multi-fidelity model. Besides, only the input data mapping values are used instead of nominal mapping functions over the whole input space. This allows a more flexible mapping adequate in the case of computationally expensive mappings. Moreover, using a GP as a mapping enables a non-parametric mapping and induces uncertainty quantification on the latter, which differs from the space mapping approach that requires a deterministic parametric form of the mapping to be used. This avoids over-fitting compared to parametric mapping. Finally, this model keeps the original input space correlations, since $X^l$ is used as input for $f_l(\cdot)$.

\subsection{The input mapping GPs}
The input mappings are performed with multi-output GPs that transform the inputs from the higher-fidelity input space into the lower-fidelity input space. The input mapping GPs are conditioned on mapped nominal values of the training set. The nominal mapping is obtained based on physical insights of the relationship between the fidelities. For example, if the low-fidelity variables are a subset of the high-fidelity variables, the nominal mapping is simply the identity. However, it can be more complicated. For instance, it can map the design variables of the high-fidelity into the low-fidelity space to obtain an identical defined quantity of interest (\textit{e.g.} the volume defined by the HF variables equals to the volume defined by the mapped HF variables) this may induce computationally expensive input mappings. The proposed model is convenient in this case  \textit{i.e.} the nominal mapping is not known in the whole input space but only for the training HF data. In the case where the nominal mapping is uncertain due to lack of physical insight in the relationship between the input spaces of the different fidelities, a white kernel can be added to the covariance function in order to take into account this uncertainty.

\subsection{The Evidence Lower Bound}
As in regular DGPs and MF-DGP, the computation of the marginal likelihood of MF-DGP-EP: $$p\left(\textbf{y}^s,\ldots,\textbf{y}^1, X^s_{s-1},\ldots,X^2_{1} |X^s,\ldots, X^1\right)$$ is analytically non-tractable. Approximations are necessary to obtain a lower bound on this marginal likelihood which is then maximized to train the multi-fidelity model.
As in MF-DGP, the DGP approximations followed is the sparse variational approximation named doubly stochastic inference scheme presented in \cite{salimbeni2017doubly}. At each layer $l$, a set of inducing inputs / outputs  $(Z_l,\textbf{u}_l)$ are introduced for the fidelity GP $f_l(\cdot)$, and similarly for each mapping $H_l(\cdot)$ a set of  inducing inputs / outputs $(W_l,V_l)$ are introduced. Then, the following variational approximation is considered:
\begin{equation}
 \begin{split}
&q\left(\{\{\textbf{f}^t_l\}_{l=1}^t\}_{t=1}^s,\{\textbf{u}_l\}_{l=1}^s,\{\{H^t_l\}_{l=1}^t\}_{t=1}^{s-1},\{V_l\}_{l=1}^{s-1}\right)\\&=\prod_{t=1}^s \prod_{l=1}^{t-1} \left[p(\textbf{f}^t_l|\textbf{u}_l;\{\textbf{f}^t_{l-1},H^t_l\},Z_{l-1})\right.\left.p\left(H^t_l|V_l;H^t_{l+1},W_{l+1}\right) \right]
\\&  \times \prod _{t=1}^s p(\textbf{f}^t_t|u_t;\{\textbf{f}^t_{t-1},X^t\},Z_{t-1}) \times\prod_{l=1}^s q(\textbf{u}_l) \times \prod_{l=1}^{s-1} q(V_l) 
 \end{split}
 \end{equation}
where $q(\cdot)$ is the variational distribution of the latent variables. Based on this approximation and using the Jensen inequality, a variational lower bound on the marginal likelihood is obtained: 
\begin{equation}
\begin{split}
\mathcal{L}=&\sum_{t=1}^s \sum_{i=1}^{n_t} \mathbb{E}_{q(f_t^{(i),t})} \left[\log p\left(y^{(i),t}|f_t^{(i),t}\right)\right] +\\&\sum_{t=1}^{s-1} \sum_{i=1}^{n_t} \mathbb{E}_{q(H_t^{(i),t+1})} \left[\log p\left(X^{(i),t+1}|H_t^{(i),t+1}\right)\right] - \\&  \sum_{l=1}^s KL\left[q\left(\textbf{u}_l\right)||p\left(\textbf{u}_l;Z_{l-1}\right)\right]-\\& \sum_{l=1}^{s-1} KL\left[q\left(V_l\right)||p\left(V_l;Z_{l+1}\right)\right]
\end{split}
 \end{equation}
 where $KL[\cdot||\cdot]$ corresponds to Kulblack-Leibler divergence.
\subsection{Training of MF-DGP-EM}
The training of MF-DGP-EM comes back to the maximization of the ELBO. This maximization is performed with respect to the hyperparameters of the fidelities GPs $\{\theta\}_{l=1}^s$, the hyperparameters of the input mapping multi-output GPs $\{\theta_{\text{map}}\}_{l=1}^{s-1}$, the induced inputs $\{Z_l\}_{l=1}^s, \{W_l\}_{l=1}^{s-1}$  and also the variational distributions $\{q(\textbf{u}_l)\}_{l=1}^s,\{q(V_l)\}_{l=1}^{s-1} $. 
\subsubsection{The variational variables}
Including the variational distributions makes the parameter space not Euclidian, hence, the ordinary gradient is not a suitable direction to follow. In fact, the variational distribution parameter space has a Riemannian structure defined by the Fisher information \cite{amari1998natural}. In this case, the natural gradient which comes back to the ordinary gradient rescaled by the inverse Fisher information matrix is the steepest descent direction.  Natural gradients were used in the case of conjugate variational inference in GP \cite{hensman2013gaussian} and also in the non-conjugate case \cite{salimbeni2018natural} where an efficient computation has been proposed. The natural gradient has been used for DGPs in \cite{hebbal2019bayesian}. This approach is used for MF-DGP-EM.  Specifically, the optimization procedure consists of a loop between an optimization step using a stochastic ordinary gradient (Adam Optimizer \cite{kingma2014adam}) with respect to the Euclidian space parameters and an optimization step using the natural gradient with respect to all the variational distributions.
\subsubsection{The induced inputs}
One of the major difficulties in MF-DGP is the optimization of the inducing inputs  $\{Z_l\}_{l=1}^s$. In \cite{cutajar2019deep}, the inducing inputs were arbitrary fixed and not optimized. In fact, except for the first layer, the inducing inputs in MF-DGP do not play the same role as in classic DGPs, where they are defined in the original input space. Specifically, the input space of the inner layers of the MF-DGP is augmented with the output of the previous layer, inducing a non-linear dependence between the $d$ first components and the $d+1$ component of each element in this augmented input space. Hence, freely optimizing $Z_l$ (with $2\leq l\leq s$) as vectors with independent components is no longer suitable. 

To overcome this issue in MF-DGP-EM, $\{Z_l\}_{l=2}^s$ are constrained as follows: 

\begin{equation}
Z_{l} = \left[Z_{l,1:d},f^*_{l-1}\left(H^*_{l-1}(Z_{l,1:d})\right)\right] ; \forall 2\leq l\leq s
\end{equation}
where $f^*_{l-1}(\cdot)$ corresponds to the mean prediction at the previous layer and $H^*_{l-1}(\cdot)$ to the mean mapped value into the lower-fidelity. This constraint keeps a dependency between $Z_{l,d+1}$ and $Z_{l,1:d}$, allowing to remove $Z_{l,d+1}$ from the expression of the ELBO. Hence the optimization is done with respect to $Z_{l,1:d}$ instead of $Z_{l}$.

\subsection{Prediction}
 The prediction of a test data $X^{*,t_0}$ belonging to the input space of fidelity $t$ using the two-level MF-DGP-EM is a two-step process. First, the test data $X^{*,t_0}$ are propagated through the first level of the MF-DGP-EM allowing the projection of the test data on the lower-fidelity inputs spaces to obtain $H^{*,t_0}_{t_0-1},\ldots,H^{*,t_0}_1$. Then, propagation through the second level is carried out to propagate the evaluation at the different fidelities. Hence, a prediction of $X^{*,t_0}$ with fidelity $t$ is:
 \begin{equation}
 q(\textbf{f}^{*}_t)= \frac{1}{k}\sum_{j=1}^k q\left(\textbf{f}_t^{j,*}|q(\textbf{u}_t);\{\textbf{f}_{t-1}^{j,*},H^{j,*,t_0}_{t}\},Z_{t-1}\right)
\end{equation}  
where $k$ is the number of propagated samples.
 \section{Experimentations}
To evaluate the performance of the proposed model MF-DGP-EM, experimentations are carried out in this Section. Firstly, analytical test problems are considered. The first analytical test case is an illustrative example to compare the different approaches and also to point out the efficiency of MF-DGP-EM on problems where classical fixed input space parametrization approaches are used (MF-DGP). The two remaining analytical test problems address the case where different dimensions and parametrizations are considered for each input space. Two physical test problems are also presented: a structural multi-fidelity problem and an aerodynamic multi-fidelity problem. The prediction accuracy is assessed using the R squared metric (R2) and the Root Mean Square Error (RMSE). The test Mean Negative Log-Likelihood (MNLL) metric is used to evaluate the uncertainty quantification on the prediction which is important for the trade-off exploration-exploitation for adaptive design of experiments and optimization (see Appendix A). For experimentations with multiple repetitions the average of a metric and its standard deviation are respectively noted $\overline{\text{metric}}$ and $std_{\text{metric}}$. The experimental setup is presented in Appendix B and the tables of the numerical results of each problem are displayed in Appendix C.
\subsection{Analytical problems}
\subsubsection{Illustrative test problem}
For this toy problem, the non-linear multi-fidelity problem proposed in \cite{perdikaris2017nonlinear} is used. The high-fidelity function $f_{hf}(\cdot)$ is defined as a function of the low-fidelity function $f_{lf}(\cdot)$: 
\begin{equation}
f_{hf}(x_1) = x_1  \exp{\left(f_{lf}(2x_1 - 0.2)\right)} - 1
\label{eq:non_l}
\end{equation}
where $f_{lf}$ is:
\begin{equation}
f_{lf}(x_1) = \cos(15 x_1)
\end{equation}
This multi-fidelity problem has been used previously in \cite{perdikaris2017nonlinear} and in \cite{cutajar2019deep} in the context of multi-fidelity modeling with the same input variable parametrization. However, one can argue that this problem can be interpreted as a multi-fidelity problem with different input space parametrizations. In fact, based on Eq.~\ref{eq:non_l} the nominal mapping $g_0(\cdot)$ between the two input spaces can be defined as:
\begin{equation}
g_0(x_{hf}) = 2x_{hf} - 0.2
\end{equation}

This nominal mapping is compared to the IMC mapping. The IMC approach is used to obtain a calibrated mapping that tries to minimize the distance between the outputs of the two fidelities according to Eq.~\ref{eq:IMC}. For this problem, a linear parametric mapping is considered for the IMC. Moreover, 14 HF training data points are sampled using a Latin Hypercube Sampling (LHS). The obtained mapping by IMC $g(\cdot)$ is the following:
\begin{equation}
g(x_{hf}) = 0.0715x_{hf} + 0.65924
\end{equation}

\begin{figure*}
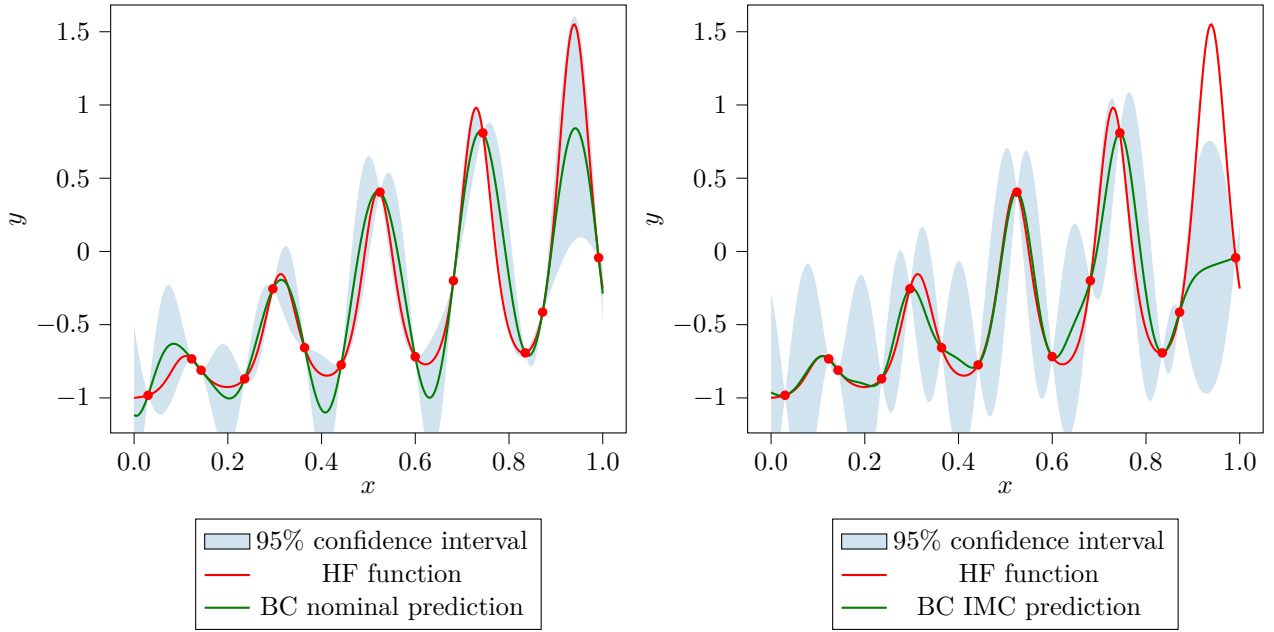

\begin{minipage}[c]{0.48\linewidth}
\input{Images/BC_prediction.tex}
\end{minipage}
\hfill
\begin{minipage}[c]{0.48\linewidth}
\input{Images/IMC_prediction.tex}
\end{minipage}
\caption{The prediction and uncertainty obtained by a bias correction approach using nominal mapping values (left) and by a bias correction approach using IMC values (right). 
BC nominal metrics: $R^2$ : 0.857, RMSE: 0.205, MNLL: -0.842.
BC IMC metrics: $R^2$ : 0.432, RMSE: 0.375, MNLL: -0.318}
\label{IMC_BC_comparison_modeling}
\end{figure*}

The IMC is compared to the nominal mapping in Fig.~\ref{mappings}. The IMC minimizes the distance between the outputs of the HF and LF following Eq.~\ref{IMC}, however, in doing so it maps the HF input space to a small range interval in the LF input space $[0,1]\rightarrow [0.659,0.730]$. To analyze this mapping in the HF output space, Fig.~\ref{mappings_} represents the exact output of the HF, the output of the LF composed with the nominal mapping, and the output of the LF composed with the IMC. The IMC results in a quadratic mean trend of the HF observations and loses the sinusoidal feature of the LF. To analyze the repercussions of such behavior from a multi-fidelity model prediction accuracy point of view, a HF GP model prediction (Fig.\ref{GP_HF_predition}) is compared to a bias correction approach (Eq.\ref{eq:BC}) used with the nominal values (BC nominal) and a bias correction approach used with the IMC mapping (Fig.\ref{IMC_BC_comparison_modeling}). The BC IMC (RMSE: 0.375) deteriorates the prediction accuracy obtained by a GP model using only the HF data (RMSE: 0.345) because of the non-adequate projection from the HF to the LF. However, the BC nominal improves the prediction accuracy (RMSE: 0.205) since the LF encodes exactly the oscillations phase information about the HF (Fig.~\ref{mappings_}). 


\begin{figure*}[h!]
\begin{minipage}[c]{0.48\linewidth} 
\begin{tikzpicture}

\definecolor{color0}{rgb}{1,0.647058823529412,0}

\begin{axis}[
legend style={at={(0.5,-0.3)},anchor=north},
tick align=outside,
tick pos=left,
x grid style={white!69.0196078431373!black},
xlabel={\(\displaystyle x_{hf}\)},
xmin=-0.05, xmax=1.05,
xtick style={color=black},
xtick={-0.2,0,0.2,0.4,0.6,0.8,1,1.2},
xticklabels={−0.2,0.0,0.2,0.4,0.6,0.8,1.0,1.2},
y grid style={white!69.0196078431373!black},
ylabel={\(\displaystyle x_{lf}\)},
ymin=-0.3, ymax=1.9,
ytick style={color=black}
]
\addplot [thick, color0, dotted]
table {%
0 -0.2
0.0101010101010101 -0.17979797979798
0.0202020202020202 -0.15959595959596
0.0303030303030303 -0.139393939393939
0.0404040404040404 -0.119191919191919
0.0505050505050505 -0.098989898989899
0.0606060606060606 -0.0787878787878788
0.0707070707070707 -0.0585858585858586
0.0808080808080808 -0.0383838383838384
0.0909090909090909 -0.0181818181818182
0.101010101010101 0.00202020202020203
0.111111111111111 0.0222222222222222
0.121212121212121 0.0424242424242424
0.131313131313131 0.0626262626262626
0.141414141414141 0.0828282828282829
0.151515151515152 0.103030303030303
0.161616161616162 0.123232323232323
0.171717171717172 0.143434343434343
0.181818181818182 0.163636363636364
0.191919191919192 0.183838383838384
0.202020202020202 0.204040404040404
0.212121212121212 0.224242424242424
0.222222222222222 0.244444444444444
0.232323232323232 0.264646464646465
0.242424242424242 0.284848484848485
0.252525252525253 0.305050505050505
0.262626262626263 0.325252525252525
0.272727272727273 0.345454545454546
0.282828282828283 0.365656565656566
0.292929292929293 0.385858585858586
0.303030303030303 0.406060606060606
0.313131313131313 0.426262626262626
0.323232323232323 0.446464646464647
0.333333333333333 0.466666666666667
0.343434343434343 0.486868686868687
0.353535353535354 0.507070707070707
0.363636363636364 0.527272727272727
0.373737373737374 0.547474747474747
0.383838383838384 0.567676767676768
0.393939393939394 0.587878787878788
0.404040404040404 0.608080808080808
0.414141414141414 0.628282828282828
0.424242424242424 0.648484848484848
0.434343434343434 0.668686868686869
0.444444444444444 0.688888888888889
0.454545454545455 0.709090909090909
0.464646464646465 0.729292929292929
0.474747474747475 0.74949494949495
0.484848484848485 0.76969696969697
0.494949494949495 0.78989898989899
0.505050505050505 0.81010101010101
0.515151515151515 0.830303030303031
0.525252525252525 0.850505050505051
0.535353535353535 0.870707070707071
0.545454545454546 0.890909090909091
0.555555555555556 0.911111111111111
0.565656565656566 0.931313131313132
0.575757575757576 0.951515151515152
0.585858585858586 0.971717171717172
0.595959595959596 0.991919191919192
0.606060606060606 1.01212121212121
0.616161616161616 1.03232323232323
0.626262626262626 1.05252525252525
0.636363636363636 1.07272727272727
0.646464646464647 1.09292929292929
0.656565656565657 1.11313131313131
0.666666666666667 1.13333333333333
0.676767676767677 1.15353535353535
0.686868686868687 1.17373737373737
0.696969696969697 1.19393939393939
0.707070707070707 1.21414141414141
0.717171717171717 1.23434343434343
0.727272727272727 1.25454545454545
0.737373737373737 1.27474747474747
0.747474747474748 1.2949494949495
0.757575757575758 1.31515151515152
0.767676767676768 1.33535353535354
0.777777777777778 1.35555555555556
0.787878787878788 1.37575757575758
0.797979797979798 1.3959595959596
0.808080808080808 1.41616161616162
0.818181818181818 1.43636363636364
0.828282828282828 1.45656565656566
0.838383838383838 1.47676767676768
0.848484848484849 1.4969696969697
0.858585858585859 1.51717171717172
0.868686868686869 1.53737373737374
0.878787878787879 1.55757575757576
0.888888888888889 1.57777777777778
0.898989898989899 1.5979797979798
0.909090909090909 1.61818181818182
0.919191919191919 1.63838383838384
0.929292929292929 1.65858585858586
0.939393939393939 1.67878787878788
0.94949494949495 1.6989898989899
0.95959595959596 1.71919191919192
0.96969696969697 1.73939393939394
0.97979797979798 1.75959595959596
0.98989898989899 1.77979797979798
1 1.8
};
\addlegendentry{Nominal mapping}
\addplot [thick, green!50.1960784313725!black, dashed]
table {%
0 0.656444457923897
0.0101010101010101 0.657143797808958
0.0202020202020202 0.657843137694019
0.0303030303030303 0.658542477579079
0.0404040404040404 0.65924181746414
0.0505050505050505 0.659941157349201
0.0606060606060606 0.660640497234262
0.0707070707070707 0.661339837119323
0.0808080808080808 0.662039177004384
0.0909090909090909 0.662738516889445
0.101010101010101 0.663437856774506
0.111111111111111 0.664137196659567
0.121212121212121 0.664836536544628
0.131313131313131 0.665535876429689
0.141414141414141 0.66623521631475
0.151515151515152 0.666934556199811
0.161616161616162 0.667633896084872
0.171717171717172 0.668333235969933
0.181818181818182 0.669032575854994
0.191919191919192 0.669731915740054
0.202020202020202 0.670431255625115
0.212121212121212 0.671130595510176
0.222222222222222 0.671829935395237
0.232323232323232 0.672529275280298
0.242424242424242 0.673228615165359
0.252525252525253 0.67392795505042
0.262626262626263 0.674627294935481
0.272727272727273 0.675326634820542
0.282828282828283 0.676025974705603
0.292929292929293 0.676725314590664
0.303030303030303 0.677424654475725
0.313131313131313 0.678123994360786
0.323232323232323 0.678823334245847
0.333333333333333 0.679522674130908
0.343434343434343 0.680222014015968
0.353535353535354 0.680921353901029
0.363636363636364 0.68162069378609
0.373737373737374 0.682320033671151
0.383838383838384 0.683019373556212
0.393939393939394 0.683718713441273
0.404040404040404 0.684418053326334
0.414141414141414 0.685117393211395
0.424242424242424 0.685816733096456
0.434343434343434 0.686516072981517
0.444444444444444 0.687215412866578
0.454545454545455 0.687914752751639
0.464646464646465 0.6886140926367
0.474747474747475 0.68931343252176
0.484848484848485 0.690012772406821
0.494949494949495 0.690712112291882
0.505050505050505 0.691411452176943
0.515151515151515 0.692110792062004
0.525252525252525 0.692810131947065
0.535353535353535 0.693509471832126
0.545454545454546 0.694208811717187
0.555555555555556 0.694908151602248
0.565656565656566 0.695607491487309
0.575757575757576 0.69630683137237
0.585858585858586 0.697006171257431
0.595959595959596 0.697705511142492
0.606060606060606 0.698404851027553
0.616161616161616 0.699104190912614
0.626262626262626 0.699803530797674
0.636363636363636 0.700502870682735
0.646464646464647 0.701202210567796
0.656565656565657 0.701901550452857
0.666666666666667 0.702600890337918
0.676767676767677 0.703300230222979
0.686868686868687 0.70399957010804
0.696969696969697 0.704698909993101
0.707070707070707 0.705398249878162
0.717171717171717 0.706097589763223
0.727272727272727 0.706796929648284
0.737373737373737 0.707496269533345
0.747474747474748 0.708195609418406
0.757575757575758 0.708894949303467
0.767676767676768 0.709594289188528
0.777777777777778 0.710293629073589
0.787878787878788 0.710992968958649
0.797979797979798 0.71169230884371
0.808080808080808 0.712391648728771
0.818181818181818 0.713090988613832
0.828282828282828 0.713790328498893
0.838383838383838 0.714489668383954
0.848484848484849 0.715189008269015
0.858585858585859 0.715888348154076
0.868686868686869 0.716587688039137
0.878787878787879 0.717287027924198
0.888888888888889 0.717986367809259
0.898989898989899 0.71868570769432
0.909090909090909 0.719385047579381
0.919191919191919 0.720084387464441
0.929292929292929 0.720783727349502
0.939393939393939 0.721483067234563
0.94949494949495 0.722182407119624
0.95959595959596 0.722881747004685
0.96969696969697 0.723581086889746
0.97979797979798 0.724280426774807
0.98989898989899 0.724979766659868
1 0.725679106544929
};
\addlegendentry{IMC mapping}
\addplot [semithick, red, mark=*, mark size=1.5, mark options={solid}, only marks, forget plot]
table {%
0.77083750272019 1.34167500544038
0.122513526169459 0.045027052338917
0.459827657947335 0.71965531589467
0.563698071484434 0.927396142968868
0.185911669719403 0.171823339438806
0.897717468649567 1.59543493729913
0.403278150933333 0.606556301866665
0.823492494268065 1.44698498853613
0.0392009645662375 -0.121598070867525
0.253205941642635 0.306411883285271
0.315975342809922 0.431950685619843
0.67024582277327 1.14049164554654
0.640261625750074 1.08052325150015
0.994685474163762 1.78937094832752
};
\addplot [semithick, red, mark=*, mark size=1.5, mark options={solid}, only marks]
table {%
0.77083750272019 0.709813121568643
0.122513526169459 0.664926638859563
0.459827657947335 0.688280464248113
0.563698071484434 0.695471895831475
0.185911669719403 0.669315987051469
0.897717468649567 0.718597611426812
0.403278150933333 0.684365279000306
0.823492494268065 0.713458671406604
0.0392009645662375 0.659158522931246
0.253205941642635 0.673975082322282
0.315975342809922 0.678320899756252
0.67024582277327 0.702848691953319
0.640261625750074 0.700772746608234
0.994685474163762 0.72531115721607
};
\addlegendentry{HF training points}
\end{axis}

\end{tikzpicture}
\caption{Parametric linear mapping obtained by IMC compared to the nominal mapping in the input space.}
\label{mappings}
\end{minipage}
\hfill
\begin{minipage}[c]{0.48\linewidth}
\input{Images/projections_.tex}
\caption{The input mapping results in the output space. The IMC output minimizes the distance from the HF training data points, however, the correlation LF and HF is lost. }
\label{mappings_}
\end{minipage}
\\
\center

\begin{minipage}[c]{0.5\linewidth}
\input{Images/GP_HF_prediction.tex}
\caption{The prediction and uncertainty obtained by a GP model of the HF.
GP HF metrics: $R^2$ : 0.702, RMSE: 0.345, MNLL: -0.608
}
\label{GP_HF_predition}
\end{minipage}
\end{figure*}
\begin{figure*}
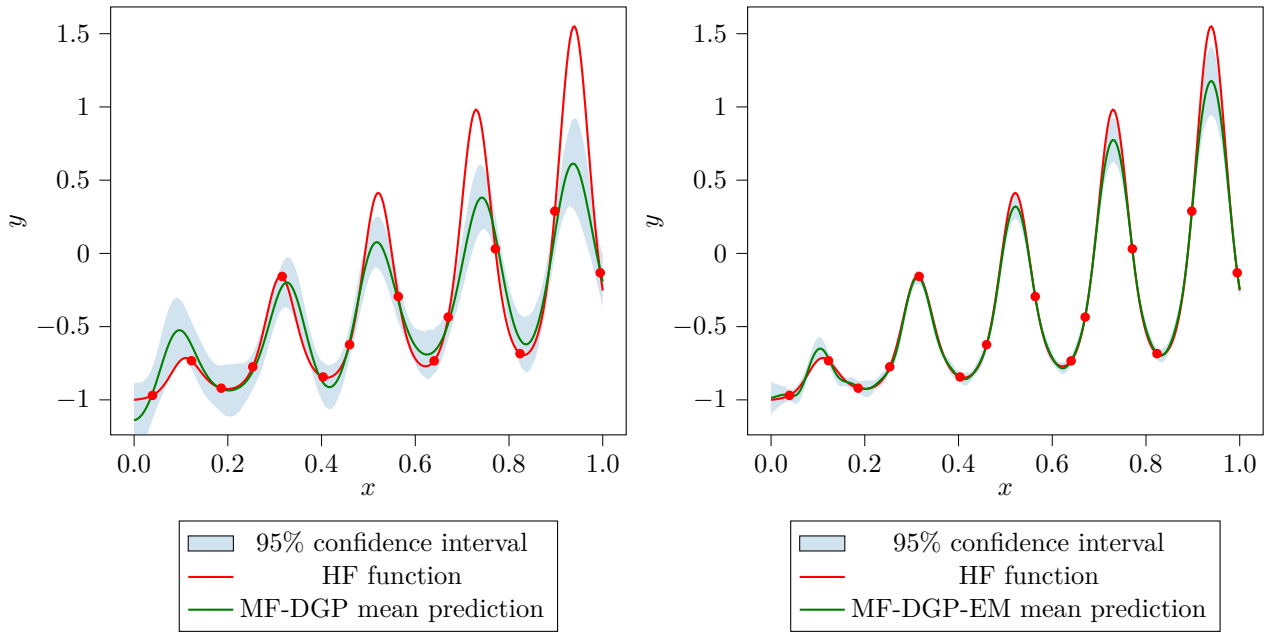

\begin{minipage}[c]{0.48\linewidth}
\input{Images/MF_DGP_prediction.tex}
\end{minipage}
\hfill
\begin{minipage}[c]{0.48\linewidth}
\input{Images/MF_DGP_EM_prediction.tex}
\end{minipage}
\caption{The prediction and uncertainty obtained by MF-DGP (left) and by MF-DGP-EM (right). MF-DGP-EM takes into account the relationship between the input spaces which improves the results. 
MF-DGP metrics: $R^2$: 0.840, RMSE: 0.255, MNLL: 0.231.
MF-DGP EM metrics: $R^2$ : 0.984, RMSE: 0.077, MNLL: -1.921.}
\label{MF_comparison_modeling}
\end{figure*}
%
MF-DGP-EM using the nominal mapped values of the HF training data is compared to the standard MF-DGP that considers the same input variable parametrization between the two problems. For these two models, the exact LF model is not used and 30 training LF points are sampled using an LHS. The MF-DGP-EM can embed the information of the input space mapping to improve the prediction accuracy and uncertainty quantification (RMSE: $0.077$, MNLL: $-1.921$) of the MF-DGP (RMSE: $0.255$, MNLL: $0.231$) as shown in Fig.~\ref{MF_comparison_modeling}. This result is interesting since the MF-DGP-EM was applied to a problem that was previously treated as a multi-fidelity problem with the same input space parametrization. Therefore, when there is some information about the input space relationship between the different fidelities, the modeling can be improved using MF-DGP-EM even in problems with the same input dimensions.

As in this case, in the next problems, the different fidelities may not share the same trend. Moreover, the LF is not considered necessarily computationally free. Hence, for comparison, the BC approach with nominal mapping is preferred to the IMC approach.

\subsubsection{Varying-input test problems}
To assess the efficiency of the proposed MF-DGP-EM, a comparison is carried out by modeling only the High-fidelity using a GP (GP-HF) and to Bias Correction approach (BC) with nominal mapping \cite{li2016integrating}. This comparison with BC is interesting since in this approach the nominal mapping functions are used to define the relationship between the fidelities, in contrast with MF-DGP-EM where only nominal mapped values of the training HF data are known and the mapping has yet to be learned. Two problems described in the following are used for this analytical comparison. \\ 
\textbf{Problem 1}: The first test case is based on the Park multi-fidelity problem \cite{xiong2013sequential}. The low-fidelity model is considered only with two variables (Eq.~\ref{Park_VD1} and Eq.~\ref{Park_VD2}). This problem depicts the case where some variables are neglected in the low-fidelity model for simplicity. The nominal mapping is naturally the identity mapping of the HF variables (Eq.~\ref{Park_VD_mapping}).\\
 The high-fidelity function is four-dimensional with an input domain $[0,1]^4$:
\begin{equation}
\begin{split}
f_{hf}(x_1,x_2,x_3,x_4) =& \frac{x_1}{2}\left(\sqrt{1+(x_2+x_3^2)\frac{x_4}{x_1^2}}-1\right)\\& + (x_1+3x_4)\exp\left(1+\sin(x_3)\right)
\end{split}
\label{Park_VD1}
\end{equation}

The low-fidelity function is two-dimensional with an input domain $[0,1]^2$:
\begin{equation}
\begin{split}
f_{lf}(x_1,x_2) =& \left(1+\frac{\sin(x_1)}{10}\right) f_{hf}(x_1,x_2,0.5,0.5)-2x_1\\&+x_2^2+0.75
\end{split}
\label{Park_VD2}
\end{equation}

The nominal mapping is a linear mapping $X^TA_0+\textbf{b}_0$ with:

\begin{equation}
A_0=\begin{bmatrix} 1& 0\\0 & 1 \\0 &0 \\0 &0\end{bmatrix}      \text{and } \textbf{b}_0= [ 0,0] 
\label{Park_VD_mapping}
\end{equation}
\textbf{Problem 2} : The second test case is a problem describing the situation where the fidelities are parameterized in different input spaces (cartesian and spherical parametrizations), in addition to different dimensionalities (Eq.~\ref{PB_spheric1} and Eq.~\ref{PB_spheric2}). 
The high-fidelity function is three-dimensional with an input domain $[0,1]^3$:
\begin{equation}
\begin{split}
f_{hf}(r,\theta,\phi) =& 3.5\left(r\cos\left(\frac{\pi}{2}\phi\right)\right) + 2.2 \left(r\sin\left(\frac{\pi}{2}\theta\right)\right)\\& +0.85\left(\left|r\cos\left(\frac{\pi}{2}\theta\right)-2r\sin\left(\frac{\pi}{2}\theta\right)\right|\right)^{2.2} \\&+ \frac{2\cos(\pi\phi)}{1 + 3r^2 + 10\theta^2}
\end{split}
\label{PB_spheric1}
\end{equation}

The low-fidelity function is two-dimensional with an input domain $[0,1]^2$:
\begin{equation}
f_{lf}(x_1,x_2) = 3x_1 + 2x_2 + 0.7(|x_1-1.7x_2|)^{2.35}
\label{PB_spheric2}
\end{equation}
The nominal mapping values are based on the transformation of the training high-fidelity points using: 
$$x = r\cos\left(\frac{\pi}{2}\phi\right)$$

$$y = r\sin\left(\frac{\pi}{2}\theta\right)$$

\begin{figure*}[t]
\begin{minipage}[c]{0.48\linewidth}
\input{Images/Park_r2_boxplot.tex}
\end{minipage}%
\hfill
\begin{minipage}[c]{0.48\linewidth}
\input{Images/Park_rmse_boxplot.tex}
\end{minipage}
\\
\center
\begin{minipage}[c]{0.48\linewidth}
\includegraphics[width=1.\linewidth]{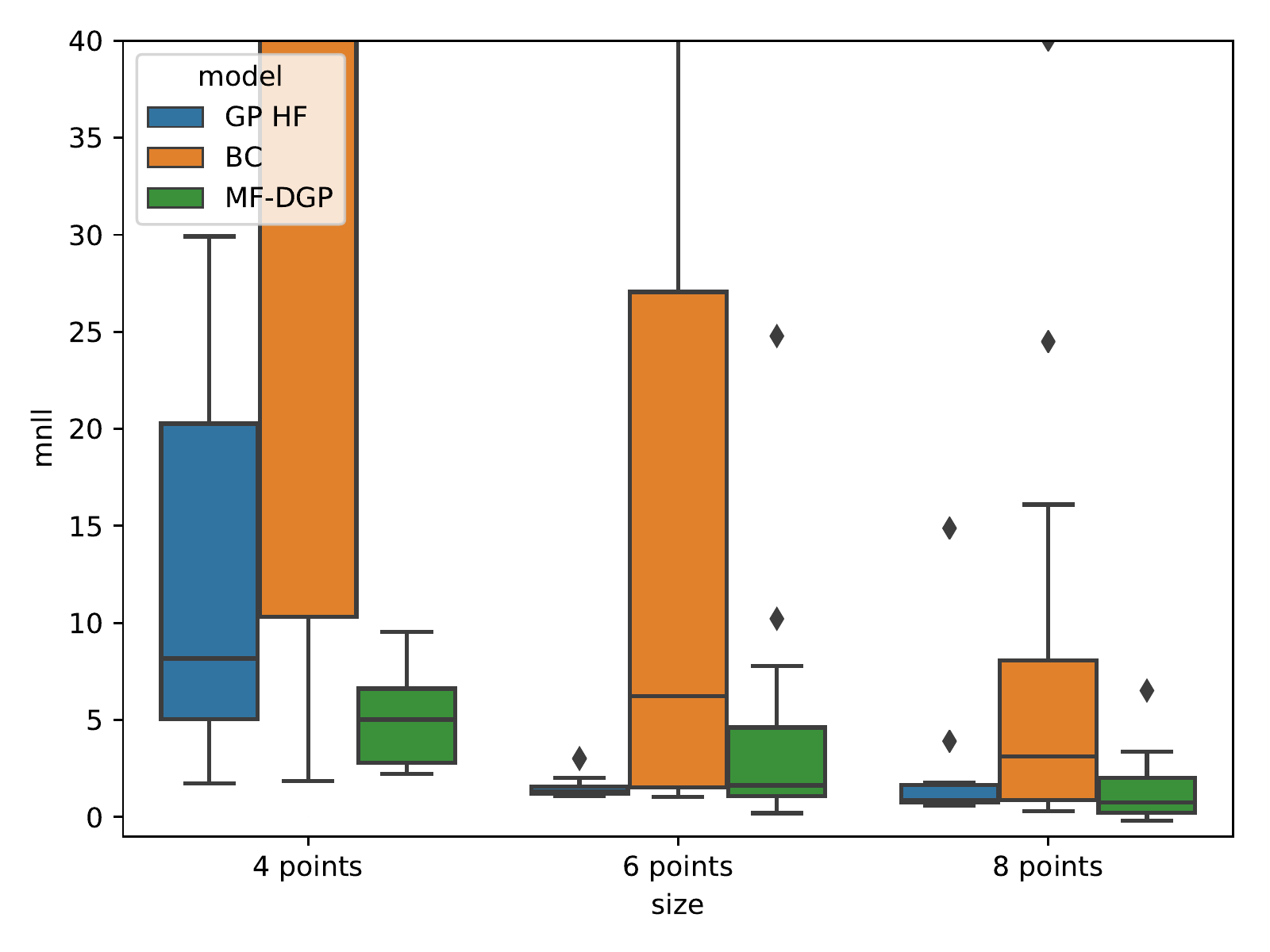}
\end{minipage}%
\caption{Performance of the different models (GP HF for the high-fidelity GP, BC for the Bias correction approach, and MF-DGP-EM for the proposed model) on \textbf{Problem 1} with 3 sizes of DoE (4, 6 and 8 data points on the HF and 30 data points on the LF) using 20 LHS repetitions.}
\label{Park_results}

\end{figure*}

\begin{figure*}[h]
\begin{minipage}[c]{0.48\linewidth}
\input{Images/Polar_r2_boxplot.tex}
\end{minipage}%
\hfill
\begin{minipage}[c]{0.48\linewidth}
\input{Images/Polar_rmse_boxplot.tex}
\end{minipage}
\\
\center
\begin{minipage}[c]{0.48\linewidth}
\includegraphics[width=1.\linewidth]{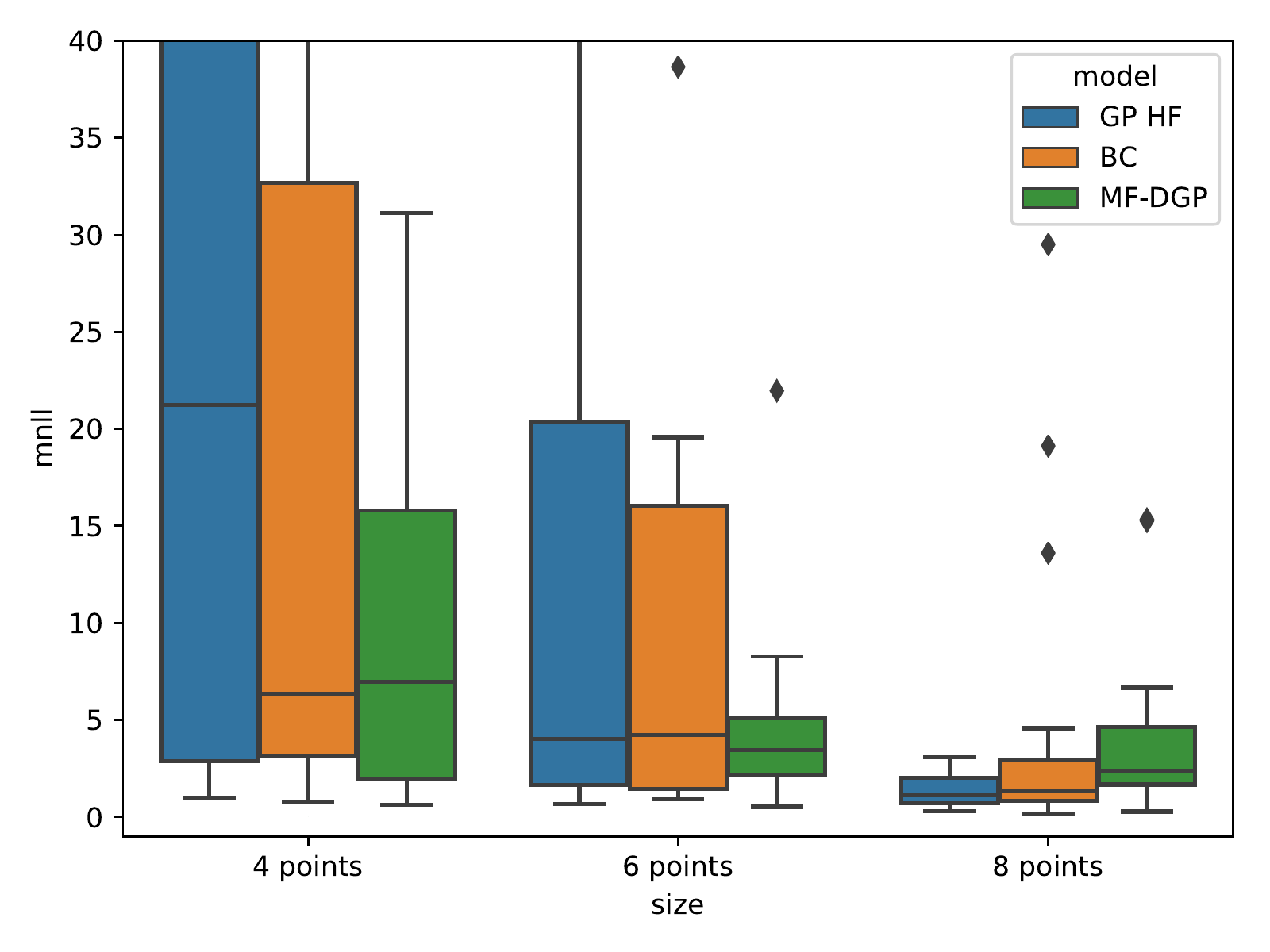}
\end{minipage}%
\caption{Performance of the different models (GP HF for the high-fidelity GP, BC for the Bias correction approach, and MF-DGP-EM for the proposed model) on \textbf{Problem 2} with 3 sizes of DoE (4, 6 and 8 data points on the HF and 30 inputs data on the LF) using 20 LHS repetitions. }
\label{Polar_results}

\end{figure*}

To assess the performance of the algorithms on different scenarios depending on the available HF information, three different sizes of the HF DoE are experimented (4, 6, and 8 HF data points). The robustness concerning the distribution of the HF data points in the input space is evaluated using 20 repetitions with different Latin Hypercube Sampling for each size of the DoE. For all the scenarios, the number of LF training data points is fixed to 30 training data  points. Fig.~\ref{Park_results} and Fig.~\ref{Polar_results} present the results obtained by the different models. On Problem 1 (Table~\ref{Resultats_park}), MF-DGP-EM is more efficient and robust to the DoE in each scenario than the other algorithms. In fact, with a DoE size of only 4 data points for HF, the MF-DGP-EM obtains better and more robust results both in terms of prediction accuracy ($\overline{RMSE}$: $1.10$, $std_{RMSE}$: $0.69$)  and uncertainty quantification ($\overline{MNLL}$: $15.75$, $std_{MNLL}$: $47.55$) compared to the BC approach ($\overline{RMSE}$: $1.83$, $std_{RMSE}$: $1.04$, $\overline{MNLL}$: $3974$, $std_{MNLL}$: $16921$ ) and the GP HF ($\overline{RMSE}$: $3.17$, $std_{RMSE}$: $1.31$, $\overline{MNLL}$: $2428$, $std_{MNLL}$: $4192$ ). The BC approach improves the prediction accuracy of the GP HF in the case where there is not enough information in the HF (4 data points). However, the relative improvement with respect to GP HF decreases when the number of HF data points crosses the threshold of 6 data points (BC $\overline{RMSE}$ for 6 HF data points: $1.25$ and for 8 HF data points: $0.93$, GP HF $\overline{RMSE}$ for 6 HF data points: $1.24$ and for 8 HF data points: $1.14$). This is not the case for the MF-DGP-EM which continues to improve the prediction accuracy even when the HF information increases (MF-DGP-EM $\overline{RMSE}$ for 6 HF data points: $0.71$ and for 8 HF data points: $0.48$). The uncertainty quantification obtained by BC is less accurate than the other approaches in the three scenarios ($\overline{MNLL}$ for 4 HF data points: $3974$, for 6 HF data points: $921$ and for 8 HF data points: $19.10$). For Problem 2 (Table~\ref{Resultats_spheric}), it is interesting to observe that in the scenario of 4 data points, the MF-DGP-EM, whilst showing improvement compared to the GP HF in term of prediction accuracy (MF-DGP $\overline{RMSE}$: $1.28$, GP HF $\overline{RMSE}$: $1.55$), it is not as good as the BC approach ($\overline{RMSE}$: $1.09$). This is due to the difficulty to learn the mapping with only 4 training data points. However, in term of uncertainty quantification, MF-DGP-EM gives the better results in the three scenarios ($\overline{MNLL}$ for 4 HF data points: $14.11$, for 6 HF data points: $4.62$ and for 8 HF data points: $3.97$) compared to either the BC approach ($\overline{MNLL}$ for 4 HF data points: $193.68$, for 6 HF data points: $93.98$ and for 8 HF data points: $4.44$) or the GP HF ($\overline{MNLL}$ for 4 HF data points: $8016$, for 6 HF data points: $468$ and for 8 HF data points: $9.14$). This is because even if there is not enough information to learn the input mapping (the case of 4 HF data points), the uncertainty quantification on this mapping is well balanced which enables the uncertainty on the prediction to be better. By increasing the HF data size (6 and 8 data points) the MF-DGP-EM learns better the mapping between the input spaces and gives also the better results in term of prediction accuracy ($\overline{RMSE}$ for 6 HF data points: $0.78$ and for 8 HF data points: $0.54$) compared to the BC approach ($\overline{RMSE}$ for 6 HF data points: $0.87$ and for 8 HF data points: $0.63$) and the GP HF approach ($\overline{RMSE}$ for 6 HF data points: $1.22$ and for 8 HF data points: $0.79$).

In conclusion of these two first experiments, the MF-DGP-EM presents generally better results in terms of prediction accuracy, uncertainty quantification, and robustness to the DoE when the mapping relationship is well learned.

\subsection{Structural problem}
\label{structural}
The first physical problem is a structural modeling problem. The objective is to model the maximum distortion criterion (also known as von Mises yield criterion) of a cantilever beam with a rectangular hole inside. This criterion expresses the needed elastic energy of distortion for the yielding of the structure to begin. The Euler-Bernoulli beam theory \cite{bauchau2009euler} is used for the low and high-fidelity models.

\begin{figure}[h]
\begin{minipage}[c]{0.48\linewidth}
\includegraphics[width=1.\linewidth]{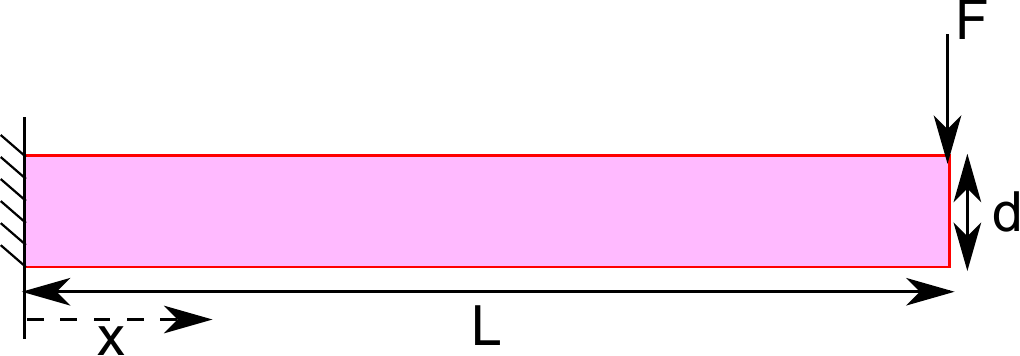}
\caption{low-fidelity beam representation. A standard solid rectangular cantilever beam characterized by its length $L$, its width $d$, and the applied force at its extremity $F$.}
\label{beam_lf}
\end{minipage}%
\hfill
\begin{minipage}[c]{0.48\linewidth}
\includegraphics[width=1.\linewidth]{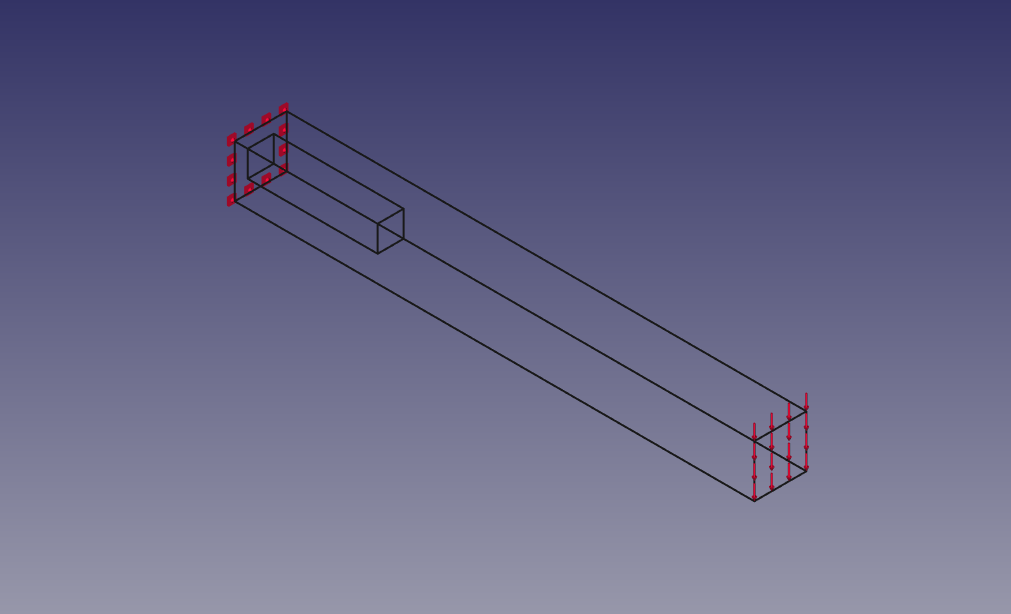}
\caption{high-fidelity beam representation. A rectangular cantilever beam with a rectangular bore along its horizontal axis.}
\label{beam_hf}
\end{minipage}%

\end{figure}

In the low-fidelity a standard solid rectangular cantilever beam (Fig.~\ref{beam_lf}) characterized by its length $L$, its width $d$, and the applied force at its extremity $F$  is considered (3 LF variables). In this case, the computation of the maximum distortion is computed analytically using the von Mises equation:
\begin{equation}
\sigma_{VM} = \sqrt{(\sigma_{ax} + \sigma_b)^2 + 3\tau_{sh}^2}
\end{equation}
where $\sigma_{ax}$ is the axial stress, $\sigma_b$ the bending stress and $\tau_{sh}$ the shear stress. For this simplified cantilever beam problem, the maximal von Mises (VM) stress is reached at the basis of the beam (meaning at $x=0$ on Fig.\ref{beam_hf}). At the basis, the axial stress is null, the shear stress is given by $\tau_{sh} = \frac{F}{l^2}$ and the bending stress is equal to $\sigma_b = \frac{6F\times L}{d^3}$.
Therefore, given the parameters $F,\; L$ and $d$, it is possible to easily estimate analytically the maximal VM within the beam.

In the high-fidelity, a rectangular cantilever beam with a rectangular bore along its horizontal axis is considered (Fig.~\ref{beam_hf}). The HF variables are the length and width of the cantilever beam, the applied force at its end, and also the width and length of the rectangular bore (5 HF variables). 

\begin{figure}[h]
\begin{minipage}[c]{0.48\linewidth}
\includegraphics[width=1.\linewidth]{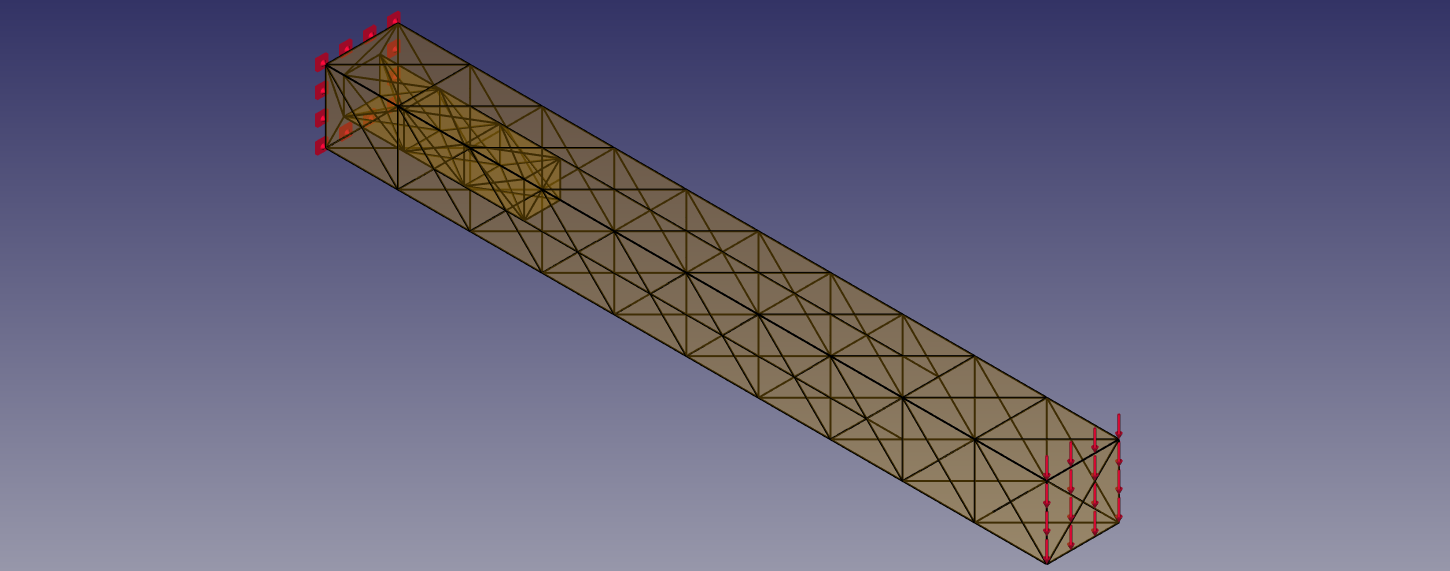}
\caption{Mesh grid used for the FE analysis of the high-fidelity cantilever beam.}
\label{mesh}
\end{minipage}%
\hfill
\begin{minipage}[c]{0.48\linewidth}
\includegraphics[width=1.\linewidth]{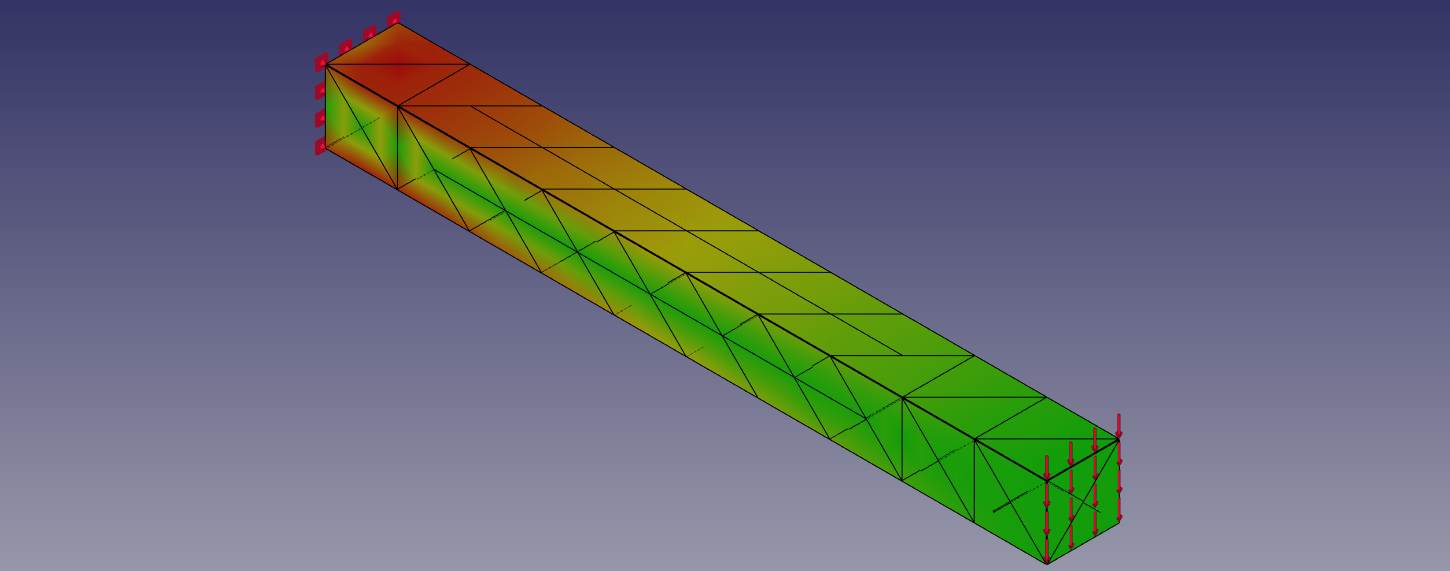}
\caption{Obtained distortion along the cantilever beam using the FE analysis.}
\label{distortion}
\end{minipage}%

\end{figure}

The maximum distortion can not be computed analytically in the case of the beam considered in the HF model. It is necessary to follow a finite element (FE) analysis approach. In this case, Caculix solver \cite{dhondt2017calculix} is used. A FE analysis can be computationally expensive according to the mesh refinement used (Fig.~\ref{mesh} and Fig.~\ref{distortion}). Hence, only a few evaluations of the HF are available. In the present case, the LF model provides an appropriate approximation of the HF model with a reduced computational cost, which makes interesting the use of multi-fidelity approaches to enrich the HF with LF information. However, the classical multi-fidelity approaches can not be used because of the difference in dimensionality between the input spaces of the HF and LF models (3 for the LF and 5 for the HF). Hence, MF-DGP-EM is used and compared to the BC approach and to using only the HF information (GP HF). Since in this case the LF design variables are included in the HF design variables, the nominal mapping is the identity with omission of 2 variables (the length and the width of the rectangular bore).
\begin{figure*}[h!]
\begin{minipage}[c]{0.48\linewidth}
\input{Images/Structure_r2_boxplot.tex}
\end{minipage}%
\hfill
\begin{minipage}[c]{0.48\linewidth}
\input{Images/Structure_rmse_boxplot.tex}
\end{minipage}
\\
\center
\begin{minipage}[c]{0.48\linewidth}
\includegraphics[width=1.\linewidth]{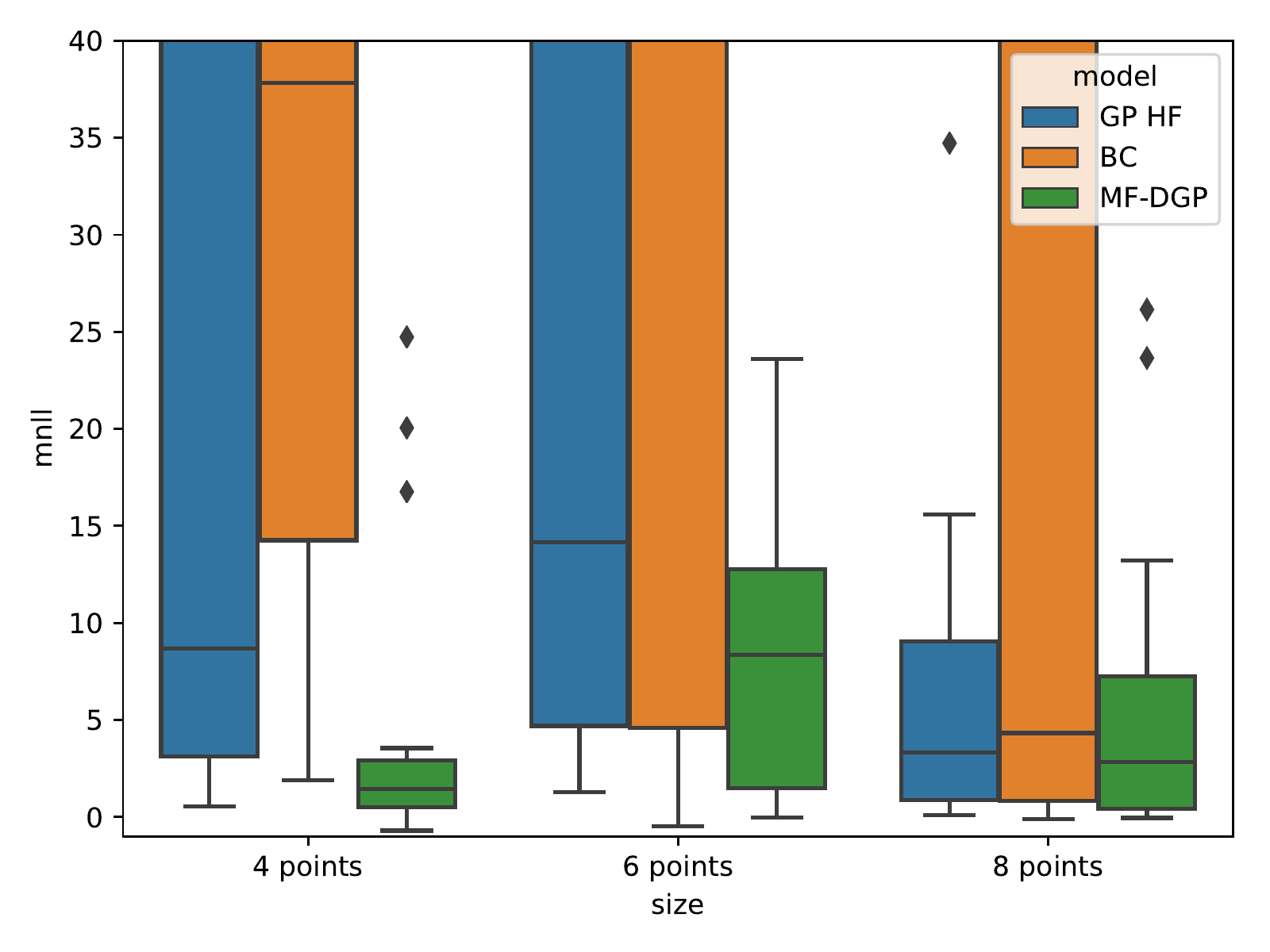}
\end{minipage}%
\caption{Performance of the different models (GP HF for the high-fidelity GP, BC for the Bias correction approach, and MF-DGP-EM for the proposed model) on the structural test problem with 3 sizes of DoE (4, 6 and 8 data points on the HF and 30 data points on the LF) using 20 LHS repetitions.}
\label{beam_results}

\end{figure*}
The performance of the models is assessed on different scenarios of the available HF information. In fact, three different sizes of the HF DoE are experimented (4, 6, and 8 data points). The robustness with respect to the distribution of the HF data points in the input space is evaluated using 20 repetitions with different Latin Hypercube Sampling for each size of the DoE. For all the scenarios the number of LF training data points is fixed to 30 training data points.

The results obtained are presented in Table~\ref{Resultats_structural} and illustrated in Fig.~\ref{beam_results}. In terms of prediction accuracy, the GP-HF is outperformed by the multi-fidelity approaches in the three scenarios which highlights the relevance of the low-fidelity model. With a DoE size of only 4 data points for HF, the BC approach outperforms the MF-DGP-EM approach in terms of prediction accuracy (BC: $\overline{RMSE}$: $0.35$, $std_{RMSE}$: $0.10$, MF-DGP-EM: $\overline{RMSE}$: $0.83$, $std_{RMSE}$: $0.4$). This can be explained by the fact that the relationship between the two fidelities is well approximated by a linear function, which makes it easier for the BC approach to capture the HF with only few information. By increasing the size of the training HF data (6 and 8 data points), the MF-DGP-EM gives comparable results to the BC approach in terms of prediction accuracy (MF-DGP-EM for 6 HF data points $\overline{RMSE}$: $0.33$ and for 8 HF data points $\overline{RMSE}$: $0.23$; BC for 6 HF data points $\overline{RMSE}$: $0.31$ and for 8 HF data points $\overline{RMSE}$: $0.25$). However, as observed in the analytical test problems, one of the main advantages of the MF-DGP-EM is the quality of the uncertainty quantification. In fact, even if the prediction accuracy is not as good as the one obtained by the BC approach (case of 4 HF data points) the added uncertainty on the nominal mapping allows the MF-DGP-EM  to obtain better results in terms of uncertainty quantification (MF-DGP-EM  $\overline{MNLL}$: $4.26$; BC $\overline{MNLL}$: $11542$ in the case of 4 HF data points). The BC approach gives less accurate results in the three scenarios when it comes to uncertainty quantification ($\overline{MNLL}$ for 6 HF data points: $14866$ and for 8 HF data points: $76.9$).

\subsection{Aerodynamic problem}
\label{aerodynamic}
In this problem, the objective is to model the lift coefficient (CL) of a winged reusable launch vehicle composed of a core, two wings, and two canards\cite{brevault2020multi}. The Vortex lattice method (VLM), is used for the computation of CL using openVSP and VSPAERO \cite{gloudemans1996rapid}. It is a computational fluid dynamics numerical approach, that models lifting surfaces, using discrete vortices to compute lift and induced drag. The span of the main wings and the canards are fixed for the two fidelities and flight conditions of Mach number equal to 0.5  and angle of attack of 2 degrees are considered.

\begin{figure}[h]
\center
\includegraphics[width=0.5\linewidth]{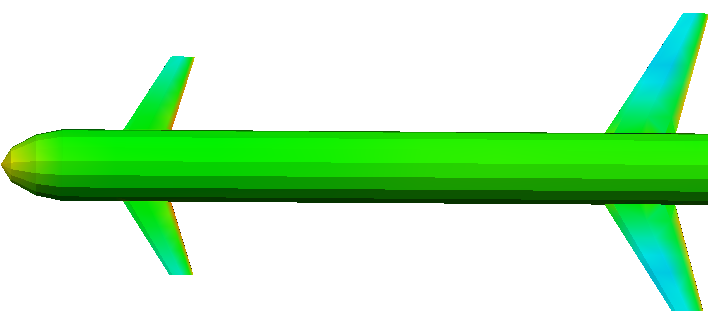}
\caption{low-fidelity winged reusable vehicle representation}
\label{LF_aero}
\center
\includegraphics[width=0.5\linewidth]{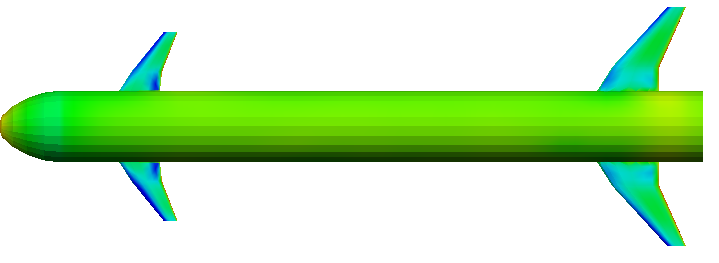}
\caption{high-fidelity winged reusable vehicle representation}
\label{HF_aero}
\end{figure}

In the low-fidelity, wings and canards with only one section are considered. The variables involved in this case are: 
\begin{itemize}
\item root chord ($RC_m$) of the main wings,
\item tip chord ($TC_m$) of the main wings,
\item sweep angle ($\beta_m$) of the main wings,
\item root chord ($RC_c$) of the canards,
\item tip chord ($TC_c$) of the canards,
\item sweep angle ($\beta_c$) of the canards.
\end{itemize}
Thus, the input space of the LF is 6-dimensional. As mentioned previously, some LF models, even though they are less computationally expensive than the HF, they are still not computationally free. This is the case in this problem where the low-fidelity configuration requires a simplified CFD analysis for the computation of CL based on VLM. 

In the high-fidelity configuration, wings and canards with two sections are considered and meshes have been densified (number of tessellated curves has been doubled). The variables involved in this case are: 
\begin{itemize}
\item root chord ($RC_m$) of the main wings,
\item tip chord ($TC_{m_1}$) of the first section of the main wings,
\item tip chord ($TC_{m_2}$) of the second section of the main wings,
\item sweep angle ($\beta_{m_1}$) of the first section of the main wings,
\item sweep angle ($\beta_{m_2}$) of the second section of the main wings,
\item relative span $\alpha_m$ of the first section of the main wings,
\item root chord ($RC_c$) of the canard,
\item tip chord ($TC_{c_1}$) of the first section of the canards,
\item tip chord ($TC_{c_2}$) of the second section of the canards,
\item sweep angle ($\beta_{c_1}$) of the first section of the canards,
\item sweep angle ($\beta_{c_2}$) of the second section of the canards,
\item relative span $\alpha_c$ of the first section of the canards,
\end{itemize}

\begin{figure*}[t]
\begin{minipage}[c]{0.48\linewidth}
\input{Images/Aero_r2_boxplot.tex}
\end{minipage}%
\hfill
\begin{minipage}[c]{0.48\linewidth}
\input{Images/Aero_rmse_boxplot.tex}
\end{minipage}
\\
\center
\begin{minipage}[c]{0.48\linewidth}
\includegraphics[width=1.\linewidth]{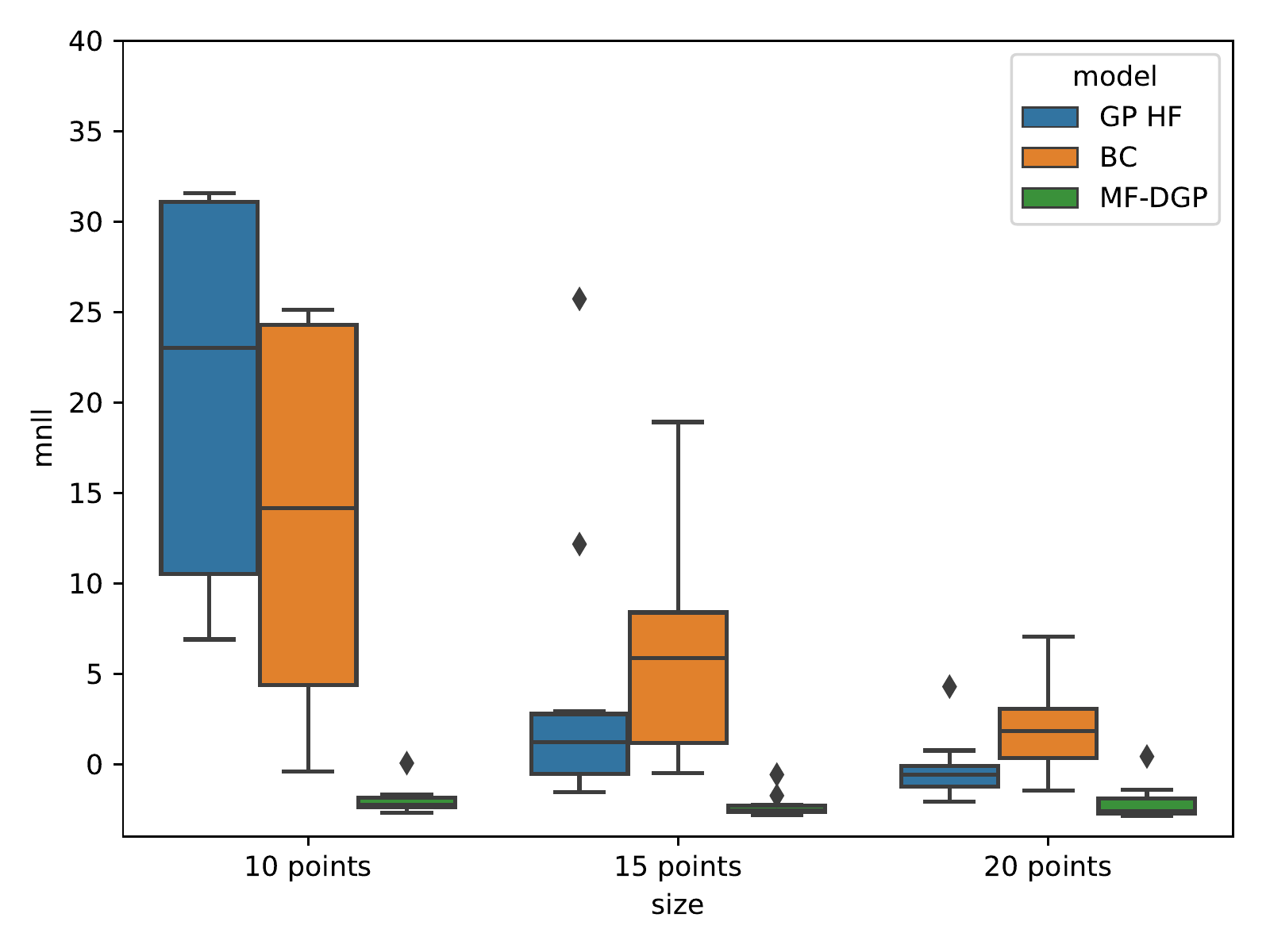}
\end{minipage}%
\caption{Performance of the different models (GP HF for the high-fidelity GP, BC for the Bias correction approach, and MF-DGP-EM for the proposed model) on the structural test problem with 3 sizes of DoE (4, 6 and 8 data points on the HF and 30 inputs data on the LF) using 20 LHS repetitions.}
\label{aero_results}
\end{figure*}

Fig.~\ref{LF_aero} and Fig.~\ref{HF_aero} illustrate the two fidelities configurations. The input space of the HF is 12-dimensional. Moreover, the mesh is refined in the HF with a doubled number of tessellated curved compared to the LF (Fig.~\ref{sections_hf_lf}). This makes the computation of CL in the HF case more complex than the LF configuration. This restrains the number of evaluations of the HF model, which makes multi-fidelity approaches interesting to enrich the HF with LF information. HF and LF models have different input space dimensions (6 for the LF and 12 for the HF). MF-DGP-EM is used and compared to the BC approach and to a GP using only the HF information (GP HF). A possible nominal mapping between the input spaces of the HF and LF is a mapping that for a set of HF design variables maps a LF design variables with the same canards and main wings surface:
\begin{equation}
\begin{split}
&RC_{bf} = RC_{hf}\\
&TC_{bf} = TC_{hf_1} + (1-\alpha_{hf}) TC_{hf_2} + (\alpha_{hf}-1) RC_{hf}\\
&\beta_{bf} =\alpha_{hf} \beta_{hf_1} + (1-\alpha_{hf}) \beta_{hf_2}
\end{split}
\end{equation}
The performance of the models is assessed on different scenarios of the HF information available. In fact, three different sizes of the HF DoE are experimented (10, 15 and 20 data points) and to evaluate the robustness with respect to the distribution of the HF data points in the input space, the experimentations have been repeated on 10 different Latin Hypercube Sampling for each size of the DoE. For all the scenarios the number of LF training data points is fixed to 120 training data points.

The obtained results are presented in Table~\ref{Resultats_aero} and illustrated in Fig.~\ref{aero_results}. MF-DGP-EM presents a better prediction accuracy and uncertainty quantification even with only 10 data points in the HF dimension ($\overline{RMSE}$: $0.023$, $\overline{MNLL}$: $-2.00$) compared to the GP HF ($\overline{RMSE}$: $0.0475$, $\overline{MNLL}$: $34.$) and the BC model ($\overline{RMSE}$: $0.0258$, $\overline{MNLL}$: $23$). Increasing the size of the HF training data allows the nominal mapping to be better learned in the case of MF-DGP-EM which enables a more significant difference between the MF-DGP-EM and the BC model in terms of prediction accuracy in the case of 15 and 20 HF training data points (MF-DGP-EM $\overline{RMSE}$ for 15 HF data points: $0.0189$ and for 20 HF data points $0.01764$; BC $\overline{RMSE}$ for 15 HF data points: $0.0248$ and for 20 HF data points $0.0222$). Some conclusions from the other experiments are also confirmed in this problem. For instance, the BC approach obtains a less accurate uncertainty quantification than the other approaches ($\overline{MNLL}$ for 15 HF data points $6.06$ and for 20 data points $1.79$) and its prediction accuracy stagnates after exceeding a threshold in the size of the training HF data (BC $\overline{RMSE}$ for 20 data points: $0.0222$, GP HF $\overline{RMSE}$ for 20 data points: $0.0207$). Also, the better uncertainty quantification of the MF-DGP-EM compared to the other approaches even when the HF available information is not enough ($10$ data points).
%
%

\subsection{Synthesis of the experiments}
These different results show the interest of using MF-DGP-EM, especially when the nominal mapping is not known for all the input space but only for the training HF data. It presents a prediction accuracy with robustness to the DoE that spares an excessive number of evaluations of the HF. Also, for the different problems, the uncertainty associated to the prediction of MF-DGP-EM  is better valued than the other approaches even in the case when the HF information is scarce. This can be explained by the uncertainty quantification on the nominal mapping of MF-DGP-EM. This makes the MF-DGP-EM more interesting to use for applications where there is a trade-off exploitation/exploration to be made such as optimization or design of experiments applications.

 \section{Conclusion and Future Works}
 Multi-fidelity problems with varying input space parametrization are common in physical and industrial applications. However, they are often addressed with models not specific to the problematic and appropriate models are scarce in the literature. In this paper, a new model for this problematic is developed. The proposed model embeds into the existing multi-fidelity Deep Gaussian Process model a mapping between the input spaces using Gaussian processes. The proposed model allows a joint optimization of the input space mapping and the multi-fidelity model, keeping the correlations in the original high-fidelity input space and allowing an uncertainty quantification of the input space mapping. 
 
 The efficiency of the proposed model has been assessed on analytical test problems and also on physical test problems. The results obtained confirm the interest of the model with a prediction accuracy and an uncertainty quantification that show robustness to the DoE that spares an excessive number of evaluations of the high-fidelity model. 
 
 The proposed model has been applied only in the case of two fidelities. However, it can be applied to more fidelities. Hence, experiments for three different fidelities with different input parametrizations may be interesting to assess the behavior of the model in more complicated configurations but may induce a computational burden during the training of the model.

The context of multi-fidelity modeling for the analysis of complex systems has been considered in this study. The natural next extension of this work is to address the multi-fidelity optimization topic with varying input space dimensions. In this perspective, this model can be coupled to Bayesian Optimization algorithms or to space mapping multi-fidelity optimization approaches.

\section*{Acknowledgments}

The work of Ali Hebbal is a funded by ONERA - The French Aerospace Lab and the University of Lille through a PhD thesis. 
This work is also part of two projects (HERACLES and MUFIN) funded by ONERA.\\
The Experiments presented in this paper were carried out using the Grid’5000
testbed, supported by a scientific interest group hosted by Inria and including CNRS, RENATER and several Universities as well as other organizations (see https://www.grid5000.fr).
\bibliographystyle{unsrt}
\bibliography{Biblio_multifi}

\section{Appendices}
\subsection{Appendix A: Metrics}
For a test set $(X^*,\textbf{y}^*)$ of size $n_{\text{test}}$ and its corresponding predicted values $\hat{\textbf{y}^*}$ and variance on the prediction $\hat{\sigma^*}^2$. The following metrics are used:

\begin{itemize}
\item \textbf{R squared}: $$ R^2 = 1-\frac{\displaystyle\sum_{i=1}^{n_{\text{test}}} \left({y^*}^{(i)} - \hat{y^*}^{(i)}\right)^2}{\displaystyle\sum_{i=1}^{n_{\text{test}}} \left({y^*}^{(i)}-\overline{y} \right)} $$
with $\overline{y}$ is the mean of the observed data.
\item \textbf{Root Mean Square Error}: $$  RMSE =  \sqrt{\frac{\displaystyle\sum_{i=1}^{n_{\text{test}}} ({y^*}^{(i)}-\hat{y^*}^{(i)})^2}{n_{\text{test}}}}                 $$
\item \textbf{Mean Negative test Log Likelihood (Gaussian case)}: $$  MNLL = -\frac{1}{n_{\text{test}}} \left(\displaystyle\sum_{i=1}^{n_{\text{test}}} \log\left( \phi\left(\frac{{y^*}^{(i)}-\hat{y^*}^{(i)}}{\hat{\sigma^*}^{(i)}} \right) \right)\right) $$ where $\phi$ denotes the Probability Density function (PDF) of the univariate Gaussian probability distribution.
\end{itemize}

\subsection{Appendix B: Experimental setup}
\begin{itemize}
\item All experiments were executed on Grid'5000 using a Tesla P100 GPU. The code is based on GPflow \cite{GPflow2017}, Doubly-Stochastic-DGP \cite{salimbeni2017doubly}, and emukit \cite{emukit2019}.
\item For all GPs, Automatic Relevance Determination (ARD) Squared Exponential (SE) kernels are used with a length-scale and variance initialized to 1. The data is scaled so the HF data have a zero mean and a variance equal to 1.
\item The Adam optimizer is set with $\beta_1=0.9$ and $\beta_2=0.99$ and a step size $\gamma^{adam}=0.003$.
\item The natural gradient step size is initialized for all layers at $\gamma^{nat}=0.01$
\item The number of training iterations for MF-DGP-EM is fixed to 28000 iterations (one iteration = Adam step + natural gradient step).
\item The mean of the variational distribution of the inducing variables for the layer $t$ is initialized at $\textbf{y}^t$, and for the input mapping GP at layer $t$ at $X^{t+1}_t$.
\item The inducing input of the fidelity GP at layer $t$ is initialized at $X^t$, and for the input mapping GP at layer $t$ it is initialized at $X^{t+1}$.
\item A Github repository featuring MF-DGP-EM will be available after the publication of the paper.  
\end{itemize}

\subsection{Appendix C: Numerical results}
\label{table_results}

\begin{table*}[h]
\centering	
\caption{Performance of the different multi-fidelity models on Problem 1 (Eqs.~\ref{Park_VD1},~\ref{Park_VD2}) using 20 repetitions with different LHS generated DoE. Three scenarios on the available HF information are experimented (4,6 and 8 inputs data on the HF). 30  train data points are used for the LF  and $1000$ test data points to compute the metrics in the HF space.}
\begin{tabular}{|c|c|c|c|c|c|c|c|}
\hline
 \multicolumn{8}{|c|}{\textbf{Analytical Problem 1}}\\
  \hline \textbf{HF DoE size} & \textbf{Algorithms} & $\overline{R^2}$ & $std_{R^2}$ & $\overline{RMSE}$ & $std_{RMSE}$ & $\overline{MNLL}$ & $std_{MNLL}$\\
\hline
  & HF model & 0.4381 & 0.4511 & 3.1673 & 1.3076 & 3974.3 & 16921.3 \\
  \cline{2-8}
 4 data points & BC model & 0.7877 & 0.3718 & 1.8324 & 1.0386 & 2428.4 & 4192.4\\
  \cline{2-8}
 & MF-DGP-EM & \textbf{0.9187 } & \textbf{0.1505}  & \textbf{1.1020}&  \textbf{0.6964 } & \textbf{15.756}  & \textbf{47.556}\\
 \hline
   & HF model & 0.9112 & 0.1046 & 1.2378 & 0.5670 & \textbf{1.5146} & \textbf{0.5407} \\
  \cline{2-8}
 6 data points & BC model & 0.9185 & 0.0398 & 1.2545 & 0.3581 & 921.27 & 2775.2\\
  \cline{2-8}
 & MF-DGP-EM & \textbf{0.9731 } & \textbf{0.0200}  & \textbf{0.7146}&  \textbf{0.2230 } & 3.8986  & 5.4270\\
 \hline
   & HF model & 0.9037 & 0.1686 & 1.1389 & 0.8453 & 19.105 & 75.129 \\
  \cline{2-8}
 8 data points & BC model & 0.9476 & 0.0489 & 0.9351 & 0.4686 & 13.875 & 33.041\\
  \cline{2-8}
 & MF-DGP-EM & \textbf{0.9874 } & \textbf{0.0093}  & \textbf{0.4784}&  \textbf{0.1803 } & \textbf{1.3614}  & \textbf{1.5949}\\
 \hline
\end{tabular}
\label{Resultats_park}
\end{table*}

\begin{table*}[h]
\centering	
\caption{Performance of the different multi-fidelity models on Problem 2 (Eqs.~\ref{PB_spheric1},~\ref{PB_spheric2}) using 20 repetitions with different LHS generated DoE. Three scenarios on the available HF information are experimented (4,6 and 8 inputs data on the HF). 30  train data points are used for the LF  and $1000$ test data points to compute the metrics in the HF space.}
\begin{tabular}{|c|c|c|c|c|c|c|c|}
\hline
 \multicolumn{8}{|c|}{\textbf{Analytical Problem 2}}\\
  \hline \textbf{HF DoE size} & \textbf{Algorithms} & $\overline{R^2}$ & $std_{R^2}$ & $\overline{RMSE}$ & $std_{RMSE}$ & $\overline{MNLL}$ & $std_{MNLL}$\\
\hline
  & HF model & 0.2549 & 0.3998 & 1.5514 & 0.4380 & 8016.6 & 31752. \\
  \cline{2-8}
 4 data points & BC model & \textbf{ 0.6248} & \textbf{0.2189 }& \textbf{1.0940} & \textbf{0.3336} & 193.68 & 732.25\\
  \cline{2-8}
 & MF-DGP-EM & 0.4509 & 0.4411 & 1.2813 & 0.5226 & \textbf{14.110} & \textbf{17.801} \\
 \hline
   & HF model & 0.4958 & 0.4079 & 1.2187 & 0.5225 & 468.17 & 1545.1 \\
  \cline{2-8}
 6 data points & BC model & 0.7412 & 0.2343 & 0.8742 & 0.3718 & 93.985 & 262.30\\
  \cline{2-8}
 & MF-DGP-EM & \textbf{0.7946 } & \textbf{0.1996}  & \textbf{0.7850}&  \textbf{0.3158} & \textbf{4.6228}  & \textbf{4.4710}\\
 \hline
   & HF model & 0.7867 & 0.2299 & 0.7959 & 0.3320 & 9.1492 & 33.817 \\
  \cline{2-8}
 8 data points & BC model & 0.8821 & 0.0431 & 0.6302 & 0.1171 & 4.4421 & 7.3884\\
  \cline{2-8}
 & MF-DGP-EM & \textbf{0.9111 } & \textbf{0.0465}  & \textbf{0.5372}&  \textbf{0.1459 } & \textbf{3.9798}  & \textbf{4.1756}\\
 \hline
\end{tabular}
\label{Resultats_spheric}
\end{table*}

\begin{table*}[h]
\centering	
\caption{Performance of the different multi-fidelity models on the structural problem (Section~\ref{structural}) using 20 repetitions with different LHS generated DoE. Three scenarios on the available HF information are experimented (4,6 and 8 inputs data on the HF). 30  train data points are used for the LF  and $1000$ test data points to compute the metrics in the HF space.}
\begin{tabular}{|c|c|c|c|c|c|c|c|}
\hline
 \multicolumn{8}{|c|}{\textbf{Structural problem}}\\
  \hline \textbf{HF DoE size} & \textbf{Algorithms} & $\overline{R^2}$ & $std_{R^2}$ & $\overline{RMSE}$ & $std_{RMSE}$ & $\overline{MNLL}$ & $std_{MNLL}$\\
\hline
  & HF model & 0.1977 & 1.1167 & 0.8751 & 0.4474 & 7668.9 & 21292. \\
  \cline{2-8}
 4 data points & BC model & \textbf{0.8702} & \textbf{0.1793} & \textbf{0.3471} & \textbf{0.1063} & 11542. & 38744.\\
  \cline{2-8}
 & MF-DGP-EM & 0.4997 & 0.2110  & 0.8309&  0.4070 & \textbf{4.2601}  & \textbf{7.0356}\\
 \hline
   & HF model & 0.6760 & 0.4672 & 0.5934 & 0.24831 & 53.055 & 117.96 \\
  \cline{2-8}
 6 data points & BC model & \textbf{0.9320} & \textbf{0.0130} & \textbf{0.3103} & \textbf{0.0814} & 14866. & 62243.\\
  \cline{2-8}
 & MF-DGP-EM & 0.9204 & 0.0402  & 0.3281&  0.1156 & \textbf{13.200}  & \textbf{18.663}\\
 \hline
   & HF model & 0.8032 & 0.2375 & 0.3895 & 0.1750 & 14.131 & 30.054 \\
  \cline{2-8}
 8 data points & BC model & 0.9179 & 0.0782 & 0.2496 & 0.0793 & 76.925 & 170.36\\
  \cline{2-8}
 & MF-DGP-EM & \textbf{0.9400} & \textbf{0.0362}  & \textbf{0.2285}&  \textbf{0.0768} & \textbf{5.7554}  & \textbf{7.3080}\\
 \hline
\end{tabular}
\label{Resultats_structural}
\end{table*}

\begin{table*}[t]
\centering	
\caption{Performance of the different multi-fidelity models on the aerodynamic problem (Section~\ref{aerodynamic}) using 20 repetitions with different LHS generated DoE. Three scenarios on the available HF information are experimented (10,15 and 20 inputs data on the HF). 120  train data points are used for the LF  and $250$ test data points to compute the metrics in the HF space.}
\begin{tabular}{|c|c|c|c|c|c|c|c|}
\hline
 \multicolumn{8}{|c|}{\textbf{Aerodynamic problem}}\\
  \hline \textbf{HF DoE size} & \textbf{Algorithms} & $\overline{R^2}$ & $std_{R^2}$ & $\overline{RMSE}$ & $std_{RMSE}$ & $\overline{MNLL}$ & $std_{MNLL}$\\
\hline
  & HF model & 0.0856 & 0.5964 & 0.0475 & 0.0154 & 33.919 & 34.519 \\
  \cline{2-8}
 10 data points & BC model & 0.7284 & \textbf{0.2160} & 0.0258 & \textbf{0.0085} & 23.232 & 25.911\\
  \cline{2-8}
 & MF-DGP-EM & \textbf{0.7646 } & 0.2796  & \textbf{0.0230}& 0.0105 & \textbf{-2.0030}  & \textbf{0.7456}\\
 \hline
   & HF model & 0.6358 & 0.2400 & 0.0300 & 0.0098 & 4.2273 & 8.0840 \\
  \cline{2-8}
 15 data points & BC model & 0.7522 & 0.1932 & 0.0248 & 0.0079 & 6.0628 & 5.6046\\
  \cline{2-8}
 & MF-DGP-EM & \textbf{0.8498} & \textbf{0.1386}  & \textbf{0.0189}&  \textbf{0.0071 } & \textbf{-2.3094}  & \textbf{0.6455}\\
 \hline
   & HF model & 0.8349 & \textbf{0.0769} & 0.0207 & \textbf{0.0043} & -0.2532 & 1.6970 \\
  \cline{2-8}
 20 data points & BC model & 0.8000 & 0.1562 & 0.0222 & 0.0072 & 1.7964 & 2.3597\\
  \cline{2-8}
 & MF-DGP-EM & \textbf{0.8685} & 0.1277  & \textbf{0.01764}&  0.0069 & \textbf{-2.144}  & \textbf{0.9646}\\
 \hline
\end{tabular}
\label{Resultats_aero}
\end{table*}

\end{document}